\pdfoutput=1

\documentclass[11pt]{article}

\usepackage{acl}

\usepackage{amssymb}
\usepackage{times}
\usepackage{latexsym}
\usepackage{graphicx}
\usepackage{amsmath}
\usepackage{amsfonts}
\usepackage{multirow}
\usepackage{booktabs}
\usepackage{stmaryrd}
\usepackage{subfigure}
\usepackage{subcaption}
\usepackage{pgfplots}
\usepackage{xspace}

\usepackage[T1]{fontenc}

\usepackage[utf8]{inputenc}

\newcommand{\petEmoji}{\includegraphics[height=1.0em,trim=0 .1em 0 0]{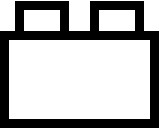}}
\newcommand{\petlWithEmoji}{PET blocks\xspace \ \petEmoji \xspace}

\usepackage{microtype}

\usepackage{inconsolata}

\title{SAPT: A Shared Attention Framework for Parameter-Efficient \\ Continual Learning of Large Language Models}

\author{Weixiang Zhao$^1$, Shilong Wang$^1$, Yulin Hu$^1$, Yanyan Zhao$^1$\thanks{\ \ Corresponding author}, Bing Qin$^1$, \\ \textbf{Xuanyu Zhang}$^2$, \textbf{Qing Yang}$^2$, \textbf{Dongliang Xu}$^2$, \textbf{Wanxiang Che}$^1$ \\
        $^1$Harbin Institute of Technology, Harbin, China\\
        $^2$Du Xiaoman (Beijing) Science Technology Co., Ltd.\\
        \texttt{\{wxzhao, yyzhao, qinb, car\}@ir.hit.edu.cn}}

\begin{document}
\maketitle
\begin{abstract}
The continual learning (CL) ability is vital for deploying large language models (LLMs) in the dynamic world. Existing methods devise the learning module to acquire task-specific knowledge with parameter-efficient tuning (PET) block and the selection module to pick out the corresponding one for the testing input, aiming at handling the challenges of catastrophic forgetting and knowledge transfer in CL. However, these methods tend to address only one of the challenges, ignoring the potential of aligning the two modules to effectively address catastrophic forgetting and knowledge transfer simultaneously. To this end, we propose a novel Shared Attention Framework (SAPT), to align the PET learning and selection via the Shared Attentive Learning \& Selection module. Extensive experiments on two CL benchmarks demonstrate the superiority of SAPT. Moreover, SAPT consistently demonstrates its superiority when we scale it to different model sizes (from 770M to 13B), different model architectures (T5 and LLaMA-2) and unseen tasks.\footnote{Our source code is available at \url{https://github.com/circle-hit/SAPT}.}
\end{abstract}

\section{Introduction}

Endowing the continual learning (CL) ability for large language models (LLMs) \citep{brown2020language,raffel2020exploring,touvron2023llama} to learn different tasks sequentially is crucial for their deployment in the real-world, which allows them to dynamically adapt to novel tasks and acquire additional knowledge \citep{luo2023investigating,zhai2023investigating,abs-2402-01364}. However, this scenario presents two significant challenges: (1) Catastrophic Forgetting (CF), referring to the loss of previously acquired knowledge when learning new tasks \citep{mccloskey1989catastrophic}, and (2) Knowledge Transfer (KT), involving the efficient utilization of knowledge from past tasks to facilitate the learning of new ones \citep{ke2022continual}.

\begin{figure}
\centering
\includegraphics[width=1\columnwidth]{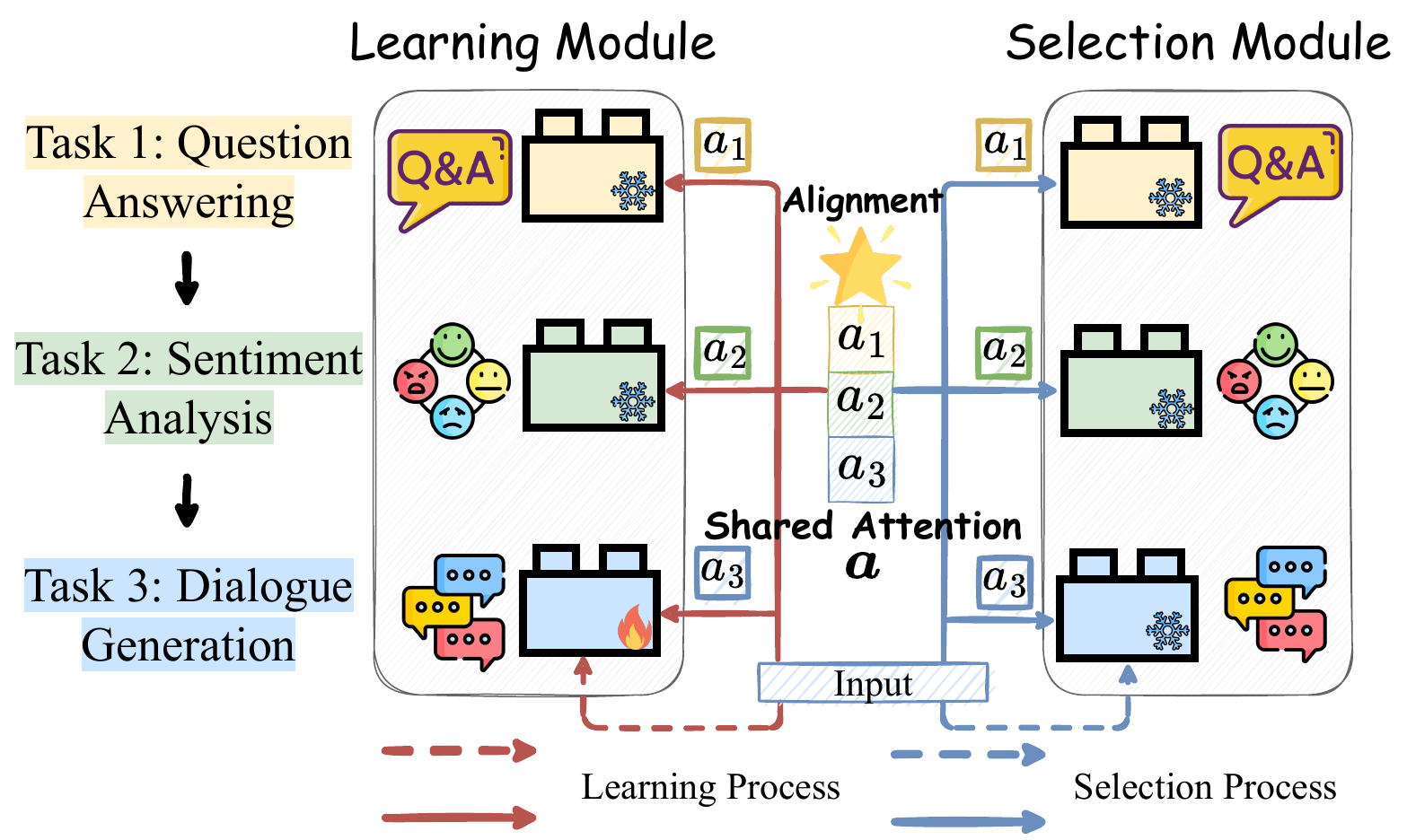}
\caption{The conceptual framework for the learning and the selection module to achieve the continual learning of large language models based on \petlWithEmoji \ when the new Dialogue Generation task arrives. Dashed lines represent the working process of existing works while solid lines are for that of our SAPT in this work.}
\label{example}
\end{figure}

Due to the heavy burden on computation resources, recent attempts study the CL of LLMs based on parameter-efficient tuning (PET) methods \citep{hu2021lora,ding2022delta}. Inspired by the parameter isolation CL methods \citep{rusu2016progressive,fernando2017pathnet}, existing methods can be conceptualized as two pivotal components working in the pipeline fashion. As shown in Figure~\ref{example} (dashed lines), when a new Dialogue Generation task arrives, a private PET block is allocated by the \emph{learning module} to acquire task-specific knowledge and then saved to the PET pool for the following \emph{selection module} to pick it out when a test sample is coming. However, the designs of each module in current works exhibit certain limitations in effectively dealing with KT and CF challenges.

\textbf{On one hand}, the design of \emph{learning module} is supposed to function to facilitate KT among different tasks. Unfortunately, for existing works, the learning of PET block is either performed seperately within each single task \citep{wang2023rehearsal}, or kept orthogonal to each other to minimize interference \citep{wang2023orthogonal}. Such isolation cuts off the potential transfer of acquired knowledge stored in the previous PET blocks and hinders them to assist the current acquisition of new knowledge.

\textbf{On the other hand}, the \emph{selection module} plays the pivotal roles in mitigating CF because only when it is capable of automatically selecting the PET block to which the current input belongs can the LLM backbone successfully accomplish the current task. However, it would make LLMs vulnerable to CF by simply implementing such selection process via the summation \citep{wang2023orthogonal} or concatenation \citep{razdaibiedina2022progressive} of all existing PET blocks or selecting them from a fixed PET pool \citep{wang2022learning}.

\textbf{More importantly}, they ignore the opportunity of aligning the two modules to address challenges of CF and KT simultaneously. The intuition is that (illustrated by solid lines in Figure \ref{example}), in order to facilitate KT in the learning of the new task, the learning module should rely on task correlations to leverage the most relevant knowledge in previous PET blocks. And such attentive process, expressed as \textbf{shared attention} in our study, could be naturally repeated by the selection module to resist CF through the combination of the corresponding PET blocks belonging to each testing input. As a result, the end-to-end alignment of these two modules is established via such shared attention.

To this end, we propose a novel \underline{\textbf{S}}hared \underline{\textbf{A}}ttention Framework for \underline{\textbf{P}}arameter-efficient con\underline{\textbf{T}}inual learning (\textbf{SAPT}) of large language models. In SAPT, the Shared Attentive Learning \& Selection Module (SALS) is devised, where each training sample is navigated to utilize the optimal combinations of existing PET blocks for completing the current task. This is achieved through an attention weight obtained via instance-level shared attention operation. Then inputs in the testing time are capable of following the same shared attention operation to reach the attention weight and pick out the appropriate PET blocks accordingly. 

However, continually updating the SALS leads to the optimal attentive combination only for the newest task, resulting in the forgetting for that of previous ones. Thus, we introduce Attentive Reflection Module (ARM) to help SALS recall what the shared attention operation of inputs from previous tasks should be originally performed with pseudo samples. And the success of ARM offers a new perspective for the utilization of generated pseudo samples instead of just blindly mixing them with samples of new tasks for multi-task training.

We conduct extensive experiments to evaluate SAPT on SuperNI \citep{wang2022super} and Long Sequence \citep{razdaibiedina2022progressive} benchmarks. State-of-the-art performance is achieved by SAPT compared with recent PET-based CL methods. Moreover, SAPT also exhibits superior performance when we scale it to different model sizes (from 770M to 13B), different model architectures, including T5 \citep{raffel2020exploring} (encoder-decoder) and LLaMA-2 \citep{touvron2023llama} (decoder-only) and previously unseen tasks.

The main contributions of this work are summarized as follows: 
\begin{itemize}
    \item We propose a novel framework SAPT, including SALS and ARM, to align the PET learning and selection process to effectively handle the CF and KT challenges simultaneously.
    \item A novel perspective for the utilization of pseudo generated samples is offered in ARM, exhibiting both improved effectiveness and efficiency than naive (generative) replay.
    \item Results of extensive experiments on the benchmark datasets demonstrate the effectiveness of SAPT to mitigate CF and facilitate KT.
\end{itemize}

\begin{figure*}[htbp]
\centering
\includegraphics[width=\textwidth]{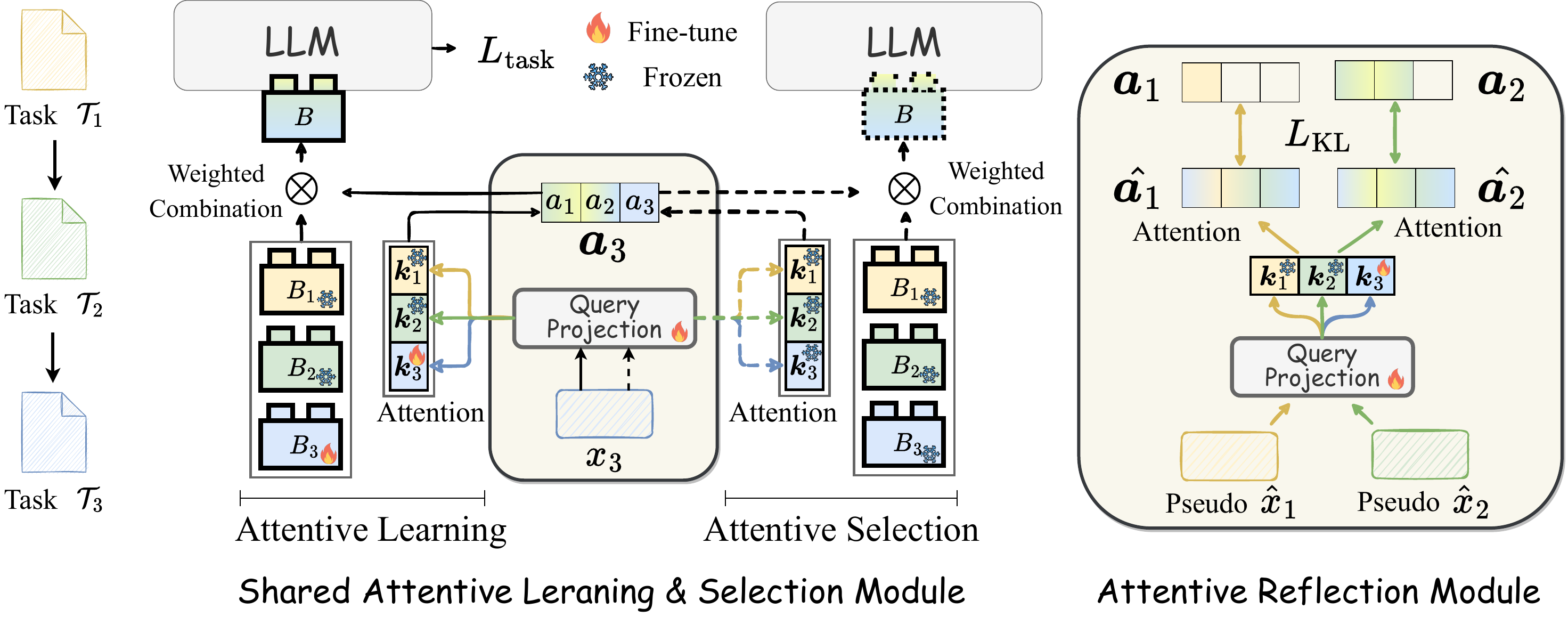}
\caption{The overall architecture of our proposed SAPT. We assume that SAPT is currently at the time step $3$ to learn the task $\mathcal{T}_3$. (1) In the SALS, as illustrated by the solid lines, the resulting attention weight $\boldsymbol{a}_3$ of task $\mathcal{T}_3$ is first obtained via the instance-level shared attention operation between the input $x_3$ and PET key vectors $\{\boldsymbol{k}_1, \boldsymbol{k}_2, \boldsymbol{k}_3\}$, to perform weighted combination of all PET blocks $\{B_1, B_2, B_3\}$ for the attentive learning of the current task $\mathcal{T}_3$. And dashed lines display the process of attentive selection, following the same process of shared attention to reach the attention weight $\boldsymbol{a}_3$ and utilizing it to handle given inputs at the testing time. (2) In the ARM, for previous tasks $\mathcal{T}_1$ and $\mathcal{T}_2$, the current attention weights of them ($\hat{\boldsymbol{a}_1}$ and $\hat{\boldsymbol{a}_2}$), are pulled back to their original states ($\boldsymbol{a}_1$ and $\boldsymbol{a}_2$), with the introduction of generated pseudo samples $\hat{x}_1$ and $\hat{x}_2$.}
\label{model}
\end{figure*}

\section{Related Works}
\subsection{Parameter-Efficient Tuning}
Recently, parameter-efficient tuning (PET) \citep{ding2022delta} has become an appealing research topic which aims at minimizing computational resources when adapting LLMs to specific tasks. Various approaches have emerged in this field, including adapter \citep{houlsby2019parameter}, prompt-based tuning \citep{lester2021power,li2021prefix}, BitFit \citep{zaken2022bitfit} and LoRA \citep{hu2021lora}. Since LoRA has exhibited superior performance compared to many mainstream PET methods, our experiments will primarily concentrate on LoRA as a representative method. To ensure a fair comparison with previous prompt-based methods, we also implement a prompt-version of SAPT.

\subsection{Continual Learning for LLMs}
\paragraph{Conventional Continual Learning (CL)} are divided into three categories. (1) \emph{Rehearsal-based methods} introduce the fixed memory to store real samples \citep{lopez2017gradient,isele2018selective} or pseudo-generative examples \citep{shin2017continual,sun2019lamol} of previous tasks. (2) \emph{Regularization-based methods} impose constraints on the loss function to penalize changes regarding the knowledge of previous tasks \citep{kirkpatrick2017overcoming,li2017learning,farajtabar2020orthogonal,WuCLLQH22,abs-2305-08698}. (3) \emph{Parameter isolation methods} dynamically expand model capacity or isolate existing model weights to mitigate interference between new and old tasks \citep{rusu2016progressive,fernando2017pathnet}.

\paragraph{Continual Learning for LLMs with PET.} Based on PET methods, current approaches for the CL of LLMs inherit the idea of parameter isolation, exhibiting a pipeline fashion to learn and select PET blocks for each task. However, most of them assume task-ids are available at testing time so that they directly use the oracle PET block of each task and just skip the selection process \citep{qin2021lfpt5,zhang2022continual,qin2023lifelong}. These lines of works simplify the problems of CL and could not be applied for real-world application of LLMs where the task-ids are unavailable. Thus, another branches of attempts focus on the more practical settings where the process of PET selection must be involved due to the unavailable task-ids during testing time. However, they are limited in effectively dealing with CF and KT challenges. For the PET learning, \citet{wang2023rehearsal} allocate private prompt for each task and \citet{wang2023orthogonal,smith2023coda} constrain the learning of PET block to keep orthogonal. They restrict the knowledge transfer among different tasks. And simply implementing the PET selection via the summation \citep{wang2023orthogonal} or concatenation \citep{razdaibiedina2022progressive} of all existing PET blocks or select them from a fixed pool \citep{wang2022learning} make LLMs vulnerable to CF.

Our proposed SAPT stands out from them in that we attempt to align the learning and selection of PET blocks so that CF and KT can be effectively addressed simultaneously.

\section{Problem Definition and Setup}

Continual learning seeks to address challenges within ongoing sequences. Formally, a sequence of tasks $\left\{\mathcal{T}_1, \ldots, \mathcal{T}_T\right\}$ arrive in a streaming fashion. Each task $\mathcal{T}_t=\left\{\left(x_t^i, y_t^i\right)\right\}_{i=1}^{n_t}$ contains a separate target dataset with the size of $n_t$. At any time step $t$, the model not only needs to adapt to the $t$-th task, but also keep performances on all previous tasks.

In this study, we delve into the more challenging and practical settings, addressing: (1) \textbf{Diverse task types}: Unlike previous approaches that merely focus on classification problems \citep{wang2023orthogonal,wang2023rehearsal}, the model would encounter a sequence of tasks encompassing various types, such as dialogue generation, information extraction, etc. (2) \textbf{Absence of task identifiers}: During the testing phase, the model confronts samples without knowing which specific task they belong to.

\section{Methodology}

\subsection{Overview of the Framework}
We propose SAPT, a novel framework for the CL of LLMs, offering an effective solution to address the challenges of catastrophic forgetting (CF) and knowledge transfer (KT) simultaneously. The overall architecture of SAPT is illustrated in Figure \ref{model}, comprising two key components: (1) Shared Attentive Learning \& Selection Module (SALS) and (2) Attentive Reflection Module (ARM). In SALS, attentive learning (solid lines) and attentive selection (dashed lines) are aligned through the shared attention operation. Then in ARM, we assist SALS in recalling the exact attentions of inputs from previous tasks with generated pseudo samples.

\subsection{Shared Attentive Learning \& Selection Module}
We devise the SALS module to align the learning and selection processes for PET blocks, where challenges of catastrophic forgetting and knowledge transfer could be effectively addressed.

\paragraph{PET Methods.} We adopt two representative PET methods, Prompt Tuning \citep{lester2021power} and LoRA \citep{hu2021lora} in SAPT. The additional trainable parameters introduced by them are referred to as PET blocks. Please refer to Appendix \ref{pet} for more details of the two PET methods.

\paragraph{Attentive Learning.}

In order to transfer the knowledge acquired from previous tasks, when the $t$-th task arrives, parameters of all previous PET blocks $\left\{B_1, B_2, \ldots, B_{t-1} \right\}$ and the current one $B_t$ are aggregated via weighted combination for the attentive learning of task $\mathcal{T}_t$. Specifically, we allocate a key vector $\boldsymbol{k}_i$ for each PET block $B_i$ ($i \in [1, t])$ and calculate instance-level input-key attentions.\footnote{\ This process is called shared attention because it will be repeated by the following attentive selection.} Such input-key attention ensures the process of attentive learning to be PET-agnostic and compatible with both prompt tuning and LoRA in SAPT.

The process of shared attention begins when the $j$-th input of the current $t$-th task passes through the embedding layer of the LLM backbone to obtain $\boldsymbol{E}_t^j$ (we will omit the superscripts $j$ for simplicity).
Since $\boldsymbol{E}_t \in \mathbb{R}^{m \times d}$ and each key vector $\boldsymbol{k}_i \in \mathbb{R}^{d}$ are of different sequence length, we first perform the max-pool operation on the length dimension of $\boldsymbol{E}_t$, and obtain $\boldsymbol{e}_{t} \in \mathbb{R}^d$. Then $\boldsymbol{e}_{t}$ is fed to a sub-network to project it as a query into the spaces of the key vectors for better feature alignment. This consists of down and up projection:
\vspace{-0.2cm}
\begin{equation}
\begin{aligned}
    \boldsymbol{h}_{t}^{\text{down}} &= \boldsymbol{W}^{\textrm{down}} (\boldsymbol{e}_{t})\\
    \boldsymbol{h}_{t}^{\text{up}} &= \boldsymbol{W}^{\textrm{up}} (\textrm{NonLinear}(\boldsymbol{h}_{t}^{\text{down}}))\\
    \boldsymbol{q}_t &= \textrm{LayerNorm} (\boldsymbol{h}_{t}^{\text{up}})\\
    \end{aligned}
\end{equation}
\label{project}where $\boldsymbol{W}^{\textrm{down}} \in \mathbb{R}^{d_p \times d}$ and $\boldsymbol{W}^{\textrm{up}} \in \mathbb{R}^{d \times d_p}$ are learnable projection parameters. Following \citet{asai2022attempt}, we use SiLU \cite{elfwing2018sigmoid} for the non-linear and apply Layer Norm \cite{ba2016layer} on $\boldsymbol{h}_{t}^{\text{up}}$ to stabilize the learning process.

Then, the attention weights $\boldsymbol{a}_t = \left\{a_1, a_2, \ldots, a_{t} \right\}$ are calculated by the product between $\boldsymbol{q}_t$ and each $\boldsymbol{k}_i$ with softmax:
\vspace{-0.2cm}
\begin{equation}
    a_{i} = \frac{\mathrm{e}^{\boldsymbol{q}_t \boldsymbol{k}_i / T}} {\sum_{i=1}^{t} \mathrm{e}^{\boldsymbol{q}_t \boldsymbol{k}_i / T}}
\end{equation}
where $T$ is a temperature factor to avoid making the attention weights over-confident and hindering the knowledge transfer. And the parameters of aggregated PET blocks can be obtained:
\vspace{-0.2cm}
\begin{equation}
    \theta_B = \sum_{i=1}^{t}a_{i} \ \theta_{B_i}
\end{equation}
\label{combine}
\vspace{-0.1cm}
where $\theta_{B_i}$ is the parameters of PET block $B_i$.

The training loss for the attentive learning of the current task $\mathcal{T}_t$ is:
\begin{equation}
L_{\text{task}} =-\!\sum_{(x_t, y_t) \in \mathcal{T}_t} \! \log P\left(y_t \mid x_t; \theta_m, \theta_B, \theta_{\text{proj}}, \theta_k \right)
\end{equation}
\label{loss}where $\theta_m, \theta_B, \theta_{\text{proj}}$ and $\theta_k$ are parameters of the LLM backbone, the aggregated PET block, the query projection layer and the set of all key vectors, respectively. And only those parameters belongs to the current $t$-th task are updated during the training, including $\theta_{B_t}$, $\theta_{\text{proj}}$  and $\theta_{k_t}$.

\paragraph{Attentive Selection.} During the inference phase, when testing data from different tasks arrives, the correct PET blocks are supposed to be automatically selected to execute the corresponding tasks. Within the preceding attentive learning, each sample has already been guided to the optimal combinations of existing PET blocks through shared attention. Thus, the attentive selection process is inherently supposed to follow the same attention operation to pick out the relevant PET blocks for the testing input accordingly. To be more specific, attentive selection involves the same computation process of Equations (1) - (3).

\paragraph{Shared Attentive Learning \& Selection.} In summary, the shared attention succeeds to align the attentive learning and selection of PET blocks, leading to the shared attentive learning \& selection that is of the same computation process and exhibiting promising insights to deal with the CF and KT challenges simultaneously.

\subsection{Attentive Reflection Module}
With the sequential training of different tasks, the query projection layer in Equation (1) is continually updated. The introduction of the Attentive Reflection Module ensures that inputs from previous tasks can still correctly perform the corresponding shared attention to identify the combination of PET blocks specific to each of them. To achieve this, we employ generative replay to constrain the projection layer with pseudo-samples. This approach ensures that no real samples are involved, thereby saving the cost associated with maintaining a fixed memory \citep{sun2019lamol,qin2021lfpt5}. 

At each time step $t$, a PET block $B_t^{\text{ref}}$ is trained to reconstruct input samples of task $\mathcal{T}_t$. For each sample (input-output pair), only the input part is generated conditioned on an initial token [Gen]. Thus, we have $\left\{B_1^{\text{ref}}, B_2^{\text{ref}}, \ldots, B_{t}^{\text{ref}} \right\}$ to obtain the generated pseudo-samples $\left\{{G}_1, {G}_2, \ldots, {G}_{t}\right\}$ (generated examples could be found in Appendix \ref{Pseudo_case}). 

To assist the query projection layer to reflect or recall the correct shared attention for samples from previous tasks at time step $t$, every instance $\hat{x}_{i}$ from ${G}_i$ is fed to the query projection layer and performs input-key attention operation following Equation (1) - (2) to obtain the current attention weight $\hat{\boldsymbol{a}_i}$. To pull $\hat{\boldsymbol{a}_i}$ to what it should originally be, we minimize a KL divergence loss:
\begin{align} \label{kl_loss}
\vspace{-0.2cm}
L_{\text{KL}} =  \sum_{i=1}^{t-1} \sum_{j=1}^{\hat{n}_i} D_{\text{KL}} (\hat{\boldsymbol{a}_i} || \boldsymbol{a}_i)
\end{align}
where $\hat{n}_i$ is the number of pseudo samples from $\mathcal{T}_i$. Here, $\boldsymbol{a}_i$ is the average attention weights of the test samples from $\mathcal{T}_i$, representing the overall attention weight of it. Notably, $\boldsymbol{a}_i$ is preserved immediately after the completion of learning $\mathcal{T}_i$, and the position of $(i, t]$ in $\boldsymbol{a}_i$ is padded with 0 when it participates the calculation in Equation (5).

It is worth to mention that our ARM exhibits both improved effectiveness and efficiency than naive (generative) replay, which is verified by the experimental results in the following Section \ref{res_and_ana}.

Finally, we jointly minimize the task loss and the KL loss in the multi-task learning fashion:
\begin{equation}
L= L_{\text{task}}+\lambda L_{\text{KL}}
\end{equation}
where $\lambda$ is a hyper-parameter that functions to balance the two parts.

\section{Experiments}
\subsection{Dataset and Evaluation Metrics}
\subsubsection{Dataset}

\paragraph{SuperNI Benchmark} \citep{wang2022super}: a benchmark of diverse NLP tasks and their expert-written instructions, enabling rigorous benchmarking of the more practical settings for the CL of LLMs. Specifically, in the types of dialogue generation, information extraction, question answering, summarization, and sentiment analysis, we select three tasks for each type, forming a sequence comprising a total of 15 tasks to evaluate various methods. For each task, 1,000 instances from the dataset are randomly sampled for training and 100 instances for validation and testing.

\paragraph{Long Sequence Benchmark} \citep{razdaibiedina2022progressive}: a continual learning benchmark of 15 classification datasets. Following \citet{razdaibiedina2022progressive,wang2023orthogonal}, we select 1,000 random samples for training each task and hold out 500 samples per class for validation and testing.

We explore two different task orders for each benchmark. Please refer to Appendix \ref{dataset} for more details about the tasks and orders.

\begin{table*}[h!]
\small
\centering
\resizebox{0.8\linewidth}{!}{
\begin{tabular}{l |c c c c |c c c c }
\toprule

\textbf{}        & \multicolumn{4}{c|}{\textbf{SuperNI Benchmark}} & \multicolumn{4}{c}{\textbf{Long Sequence Benchmark}} \\
\textbf{}        & \textbf{AP}$\uparrow$      & \textbf{F.Ra}$\downarrow$      & \textbf{FWT}$\uparrow$      & \textbf{BWT}$\uparrow$      & \textbf{AP}$\uparrow$        & \textbf{F.Ra}$\downarrow$       & \textbf{FWT}$\uparrow$       & \textbf{BWT}$\uparrow$     \\ \midrule
SeqLoRA  &  6.43   & 33.39    & -13.58   & -30.94 &   9.72   & 78.61         &  0.81   & -73.37    \\
Replay  &  35.37  &  16.92      &  -1.35   &  -15.79   & 71.28  &13.05 &  1.28          &   -12.18     \\
L2P   &  12.73      &  11.87     &   -19.14      &    -7.95   &       57.98        &      22.49      &      1.36     &     -16.63    \\
LFPT5 & 34.37 & 15.80 & -0.46 & -14.47   &   67.01      &   13.89          &    2.48       &    -12.80      \\
ProgPrompt  & 3.34   & 35.57  &  -3.29 & -33.18  &7.98   & 71.55  & -2.63      &  -66.71        \\
EPI  & -               & -            & -            & -              & 75.15              & 1.61             & -0.77             & -1.42             \\
O-LoRA  & 25.89               & 26.37            & -0.14            & -24.59              & 69.24              & 7.00             & -8.15             & -4.05             \\ \midrule
\textbf{SAPT-Prompt}     & 41.11              & 1.32              & \textbf{1.95}              & -0.65                & 79.14              & 1.68             & \textbf{3.29}             & -1.48             \\
\textbf{SAPT-LoRA}   & \textbf{51.54}               & \textbf{0.91}              & 1.88              & \textbf{-0.57}                & \textbf{82.02}              & \textbf{1.50}             & 1.86             & \textbf{-1.25}             \\ 
\bottomrule
\end{tabular}
}
\caption{The overall results on two continual learning benchmarks with T5-Large model. Performance of continual learning (AP), forgetting rate (F.Ra), forward transfer (FWT) and backward transfer (BWT) are reported after training on the last task. All results are averaged over two different orders of each benchmark.}
\label{main results}
\end{table*}

\subsubsection{Metrics}
Let $a_{i,j}$ be the testing performance (Accuracy for classification task and Rouge-L \citep{lin2004rouge} for others) on the $j$-th task after training on $i$-th task, the metrics for evaluating are:

(1) \textbf{Average Performance (AP)} \citep{chaudhry2018riemannian}. The average performance of all tasks after training on the last task, i.e., $A_{\mathcal{T}} = \frac{1}{\mathcal{T}}\sum^{\mathcal{T}}_{t=1}a_{\mathcal{T},t}$;

(2) \textbf{Forgetting Rate (F.Ra)}~\cite{chaudhry2018riemannian} measures how much knowledge has been forgotten across the first $\mathcal{T}-1$ tasks, i.e., $
    F_{\mathcal{T}} = \frac{1}{\mathcal{T}-1}\sum^{\mathcal{T}-1}_{t=1}( \textrm{max}_{k=i}^{\mathcal{T}-1} a_{k,t}-a_{\mathcal{T},t})$;
    
(3) \textbf{Forward Transfer (FWT)}~\cite{lopez2017gradient} measures how much knowledge from previous tasks transfers to a new task, i.e., $\text{FWT}_{\mathcal{T}} = \frac{1}{\mathcal{T}}\sum^{\mathcal{T}}_{t=1}(a_{t,t}-a_{0,t})$,
    where $a_{0,t}$ refers to the performance of training task $t$ individually;
    
(4) \textbf{Backward Transfer (BWT)}~\cite{ke2022continual} measures how much the learning of subsequent tasks influences the performance of a {learned} task, i.e.,
$
    \text{BWT}_{\mathcal{T}} = \frac{1}{\mathcal{T}-1}\sum^{\mathcal{T}-1}_{t=1}(a_{\mathcal{T},t}-a_{t,t})
$.

\subsection{Baselines and Comparison Models}
We evaluate SAPT against the following PET-based continual learning baseline methods: (1) \textbf{SeqLoRA}: sequentially trains the LoRA on the task orders. (2) \textbf{Replay}: replays real samples from old tasks when learning new tasks to avoid forgetting. (3) \textbf{L2P} \cite{wang2022learning}: uses the input to dynamically select and update prompts from a fixed prompt pool. (4) \textbf{LFPT5} \cite{qin2021lfpt5}: continuously trains a soft prompt for each task with generative replay and an auxiliary loss. (5) \textbf{ProgPrompt} \cite{razdaibiedina2022progressive}: sequentially concatenates previous learned prompts to the current one during the training and testing time. (6) \textbf{EPI} \cite{wang2023rehearsal}: trains prompts for each task and selects them via the distance between the input and distributions formed by labels of different classification tasks. (7) \textbf{O-LoRA} \cite{wang2023orthogonal}: learns tasks in different LoRA subspaces that are kept orthogonal to each other and sums all LoRA weights up at testing time.

\subsection{Implementation Details}
SAPT is a model- and PET-agnostic CL method that is compatible with any transformer-based generative LLM. 
In our experiments, all methods are performed with instruction tuning \cite{wei2021finetuned,ouyang2022training} to leverage the task instruction provided in the two benchmarks. To ensure a fair comparison with recent works, we implement SAPT with both prompt tuning and LoRA based on the pre-trained encoder-decoder T5-large model \cite{raffel2020exploring}. We also scale SAPT to the backbone with larger model size (up to 11B and 13B) and the decoder-only LLaMA-2 model \cite{touvron2023llama}. For the baselines, since they only report the AP metric in their original papers, we carefully re-implement them with their official codes to report metrics of F.Ra, FWT and BWT, providing a thorough insight of how existing methods deal with CF and KT. For more detailed settings, please refer to the Appendix \ref{implementation}.

\begin{table*}
\centering
\small
\begin{tabular}{l |c c c c |c c c c }
\toprule

\textbf{}        & \multicolumn{4}{c|}{\textbf{SuperNI Benchmark}} & \multicolumn{4}{c}{\textbf{Long Sequence Benchmark}} \\
\textbf{}        & \textbf{AP}$\uparrow$      & \textbf{F.Ra}$\downarrow$      & \textbf{FWT}$\uparrow$      & \textbf{BWT}$\uparrow$      & \textbf{AP}$\uparrow$        & \textbf{F.Ra}$\downarrow$       & \textbf{FWT}$\uparrow$       & \textbf{BWT}$\uparrow$     \\ \midrule
SAPT-LoRA   & \textbf{51.54}               & \textbf{0.91}              & \textbf{1.88}              & \textbf{-0.57}                & \textbf{82.02}              & \textbf{1.50}             & 1.86            & \textbf{-1.25}             \\ \midrule 
-- ARM  &  11.12                &   42.83              &     0.70            &     -40.44              &      10.18             &    78.45              &    \textbf{1.93}              &     -73.22             \\
+ Replay  & 45.41 & 7.70 & 1.26 & -6.79 & 76.93 & 6.86 & 1.21 & -6.41 \\
-- Alignment           & 45.90             & 2.98            & -2.42            & -2.55                & 77.61              & 2.83             & -3.92             & -2.48             \\
-- SA &       44.36           &    4.16             &   -2.95              &    -3.56               &     67.81           &     8.24             &    -8.60             &    -7.59           \\
\bottomrule
\end{tabular}
\caption{Results of ablation study on two benchmarks. ARM, Alignment and SA refer to the attentive reflection module, the alignment of the learning and selection in SAPT and shared attentive learning \& selection, respectively.}
\label{ablation}
\end{table*}

\section{Results and Analysis}
\label{res_and_ana}
\subsection{Overall Results}
Table \ref{main results} demonstrates the performance comparison of SAPT and recent PET-based continual learning baselines on the SuperNI and Long Sequence benchmarks. All results are averaged over the two different orders of each benchmark. Detailed results of each order and each task within a specific order are provided in Appendix \ref{overall results}.

\paragraph{Our SAPT could effectively deal with the challenges of CF and KT simultaneously.} Compared to both prompt-based methods (SAPT-Prompt v.s LFPT5/ProgPrompt/EPI) and LoRA-based methods (SAPT-LoRA v.s Replay/O-LoRA), SAPT performs better in addressing the two critical challenges, CF (highest AP and lowest F.Ra) and KT (highest FWT and BWT) when learning different tasks sequentially. Moreover, for the replay-based methods, the better performance of SAPT over Replay and LFPT5 offers a new perspective for the utilization of pseudo samples instead of just blindly mixing them with samples of new tasks for joint training. Please refer to Appendix \ref{Pseudo_analysis} for more detailed results and analysis regarding the utilization of replayed samples.

\paragraph{The alignment of learning and selection of PET is better than previous pipeline fashion.} SAPT outperforms the state-of-the-art pipeline method, EPI, which verifies the effectiveness of aligning the learning and selection with a shared attention weight. Since EPI is specifically designed for the CL of classification tasks where the selection of PET is based on the label information of each task, it can not be directly applied to the SuperNI benchmark covering various types of tasks other than classification. This manifests that SAPT is more practical to the real-world applications of LLMs. In addition, the best results of SAPT in terms of AP and F.Ra demonstrate the great potential that such attention-guided soft selection of PET are more resistant to CF, compared with previous methods of concatenation (ProgPrompt), summation (O-LoRA) and top-1 selection (EPI).

\begin{figure}
\centering
\includegraphics[width=1\columnwidth]{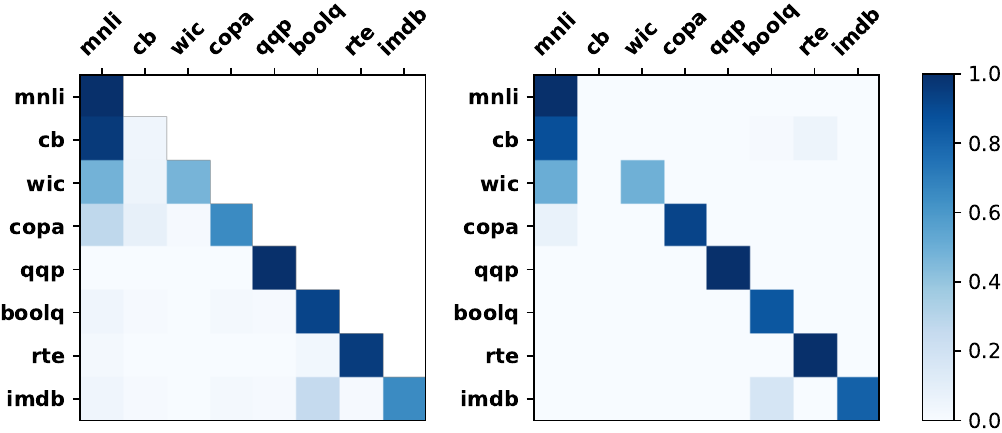}
\caption{Visualization on shared attention of SAPT-Prompt on the Long Sequence benchmark during the training for each task (left) and testing for all tasks after the training of the last task (right).}
\label{main_heat_map}
\end{figure}

\subsection{Visualization on Shared Attention}
Figure \ref{main_heat_map} displays the heat maps for shared attention during the training and testing time. We can observe that: (1) the learning and selection processes of PET blocks are exactly aligned that the two heatmaps nearly have the same layout. (2) KT do happens in the attentive learning process to assist SAPT acquire new knowledge. These further verify the effectiveness of SAPT to deal with CF and KT. Please refer to Appendix \ref{visual} for more discussions and visualization results.

\subsection{Ablation Study}
We conduct ablation studies to verify the effectiveness of different modules proposed in SAPT-LoRA. Results are shown in Table \ref{ablation}.

\paragraph{Effect of Attentive Reflection.} After removing the attentive reflection module (``-- ARM'', implemented by discarding the $L_{\textrm{KL}}$), the significant decline highlights its crucial role in assisting different input samples to recall the correct shared attention for the corresponding PET blocks they should originally combine. When replacing ARM with naive Replay (``+ Replay''), the decline of F.Ra further verifies our claim that ARM offers a more effective solution to apply pseudo samples. Please refer to Appendix \ref{Pseudo_analysis} for more detailed results and analysis regarding the efficiency of ARM module.

\paragraph{Effect of the Alignment.} We transform the alignment of PET learning and selection in SAPT into an independent format. This involves initially performing attentive learning to obtain weights that represent the combination of existing PET blocks. Subsequently, a separate PET selector is trained with these weights and generated pseudo samples. The comprehensive decline in model performance validates our claim that the learning and selection processes of PET are inherently capable of aligning together to collaborate seamlessly.

\paragraph{Effect of Shared Attentive Learning \& Selection.} Furthermore, we remove the shared attentive mechanism based on the above pipeline settings, where each PET block is learned within a single task and the selector are required to pick the most confident top-1 block for inference. The model's performance has suffered significantly, especially in terms of knowledge transfer. This demonstrates that leveraging acquired knowledge comprehensively, whether in PET learning or selection, is crucial for effectively addressing CF and KT.

\begin{figure*}
\centering
\subfigure[Results on the SuperNI benchmark]{
\includegraphics[width=0.4\textwidth]{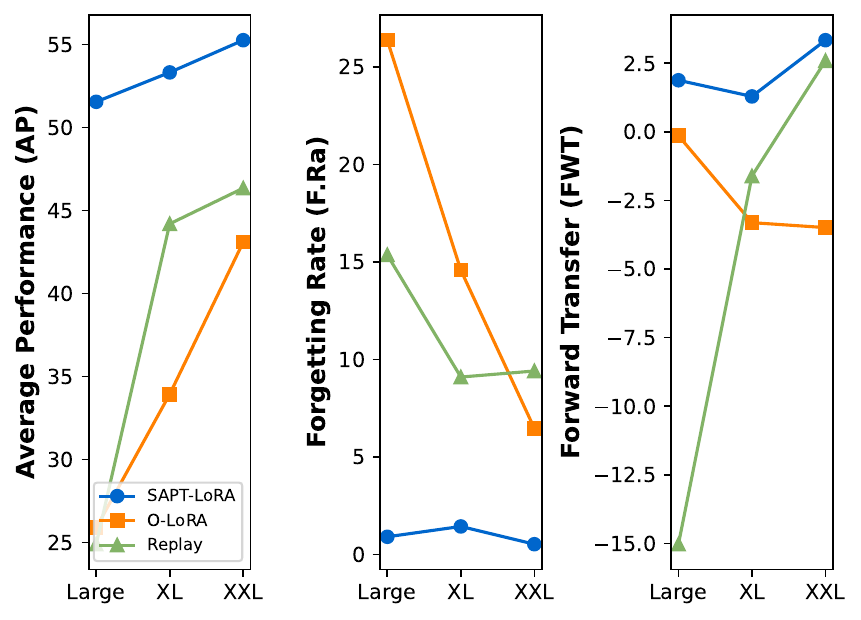}
\label{fig:subim1}
}
\subfigure[Results on the Long Sequence benchmark]{
\includegraphics[width=0.4\textwidth]{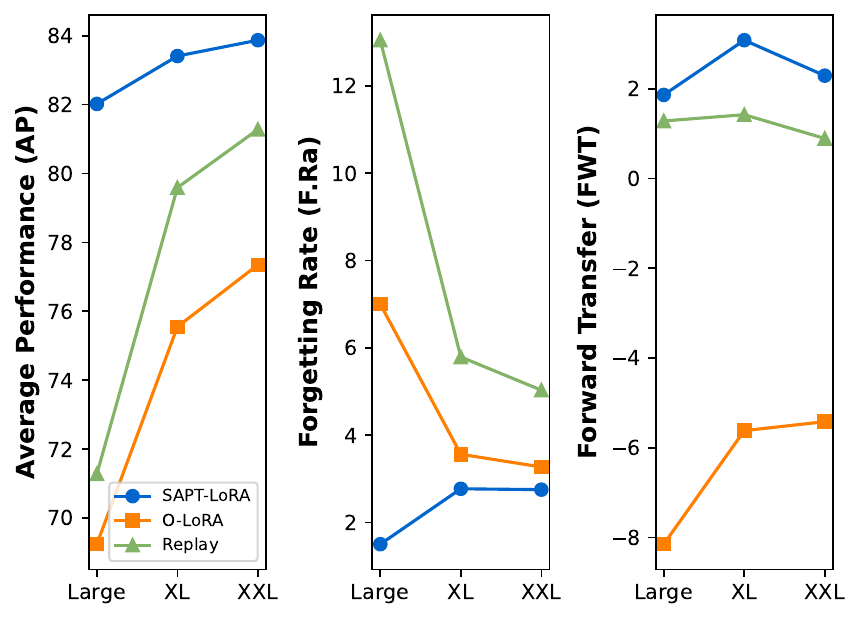}
\label{fig:subim2}
}
\caption{Performance of SAPT and baseline methods based on different size of T5-model in terms of performance of continual learning, forgetting rate and forward transfer.}
\label{t5_scale}
\end{figure*}

\subsection{Power of Scale}
\paragraph{Scale to larger backbone.} We empirically analyze how increasing the backbone T5 size affects the performance of SAPT. Figure \ref{t5_scale} displays the performance of SAPT, O-LoRA and Replay in terms of AP, F.Ra and FWT, based on the ascending backbone sizes, Large (770M), XL (3B) and XXL (11B). Overall, with the increased sizes of the backbone model, SAPT could always demonstrate superior performance over baseline models in resisting catastrophic forgetting and facilitating knowledge transfer. It is worth noting that even with the largest backbone model, O-LoRA (11B) still falls short in terms of Average Performance compared to the smallest version of SAPT-LoRA (770M). This further underscores the crucial importance of selecting the pertinent PET blocks for each input sample in real-world application scenarios.

\begin{figure}
\centering
\includegraphics[width=0.9\columnwidth]{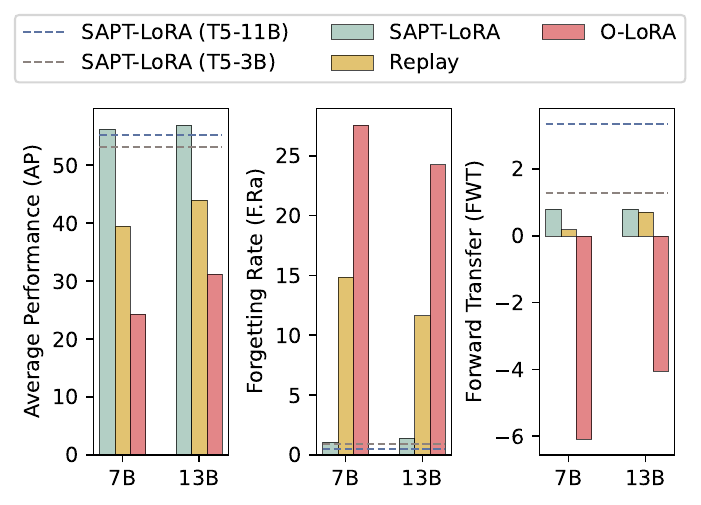}
\caption{Comparison of SAPT and baselines based on different architectures of LLM backbones, including T5 (encoder-decoder) and LLaMA-2 (decoder-only).}
\label{dif_model}
\end{figure}

\paragraph{Scale to different architectures.} The results of SAPT and baseline methods on the SuperNI Benchmark based on different sizes of T5 and LLaMA-2 are shown in Figure \ref{dif_model}. It can be observed that SAPT is capable of effectively mitigating CF and promoting KT across different model architectures. Moreover, the average performance improves with the enhancement of the model's basic capabilities (LLaMA-2 > T5). This further demonstrates the generality of our proposed SAPT. Please refer to Appendix \ref{llama_res} for more detailed results.

\paragraph{Scale to unseen tasks.} We further select 3 tasks from each one of the above task category to assess the SAPT's cross-task generalization ability. This is also a crucial dimension for evaluating CL algorithms. Table \ref{unseen} shows the results. T5-ZS represent the zero-shot approaches for task adaptation, respectively. SAPT yields the best performances, which can be attributed to its superiority in effectively combining acquired knowledge to address novel tasks. This suggests that we should actively promote knowledge transfer between different tasks during the process of CL. 

\begin{table}[]
\centering
\resizebox{\linewidth}{!}{
\begin{tabular}{
l |
c 
r 
r 
r
r
r}
\toprule
& \multicolumn{5}{c|}{\textbf{Unseen Tasks}} & \multirow{2}{*}{\textbf{Avg.}} \\
\textbf{}        & \textbf{Dialog}     & \multicolumn{1}{c}{\textbf{IE}}      & \multicolumn{1}{c}{\textbf{QA}}      & \textbf{Sum} & \multicolumn{1}{c|}{\textbf{SA}}  &   \\ \midrule
T5-ZS  &    7.49         &     6.70      &   4.28      &  12.14  &   4.54  & 
 7.03 \\
O-LoRA   &    4.39     &  9.89    &  25.38  &  8.26  &  50.41  & 19.67  \\
LFPT5   &    6.96     &  \textbf{35.32}    &  35.00  &  13.26  &  21.51  & 22.41  \\
\midrule
SAPT-LoRA   &  \textbf{11.56}     &  29.66     & \textbf{38.04}      & \textbf{13.77} & \textbf{50.62} & \textbf{28.73}         \\
\bottomrule
\end{tabular}
}
\caption{Results on unseen tasks based on the T5-Large backbone model. We report the average Rouge-L of the 3 tasks under each category.}
\label{unseen}
\end{table}

\section{Conclusion}
In this paper, we propose SAPT, a novel framework for the parameter-efficient continual learning of LLMs. In SAPT, we ingeniously align the two key processes of parameter-efficient block learning and selection through the shared attention, allowing it to effectively alleviate catastrophic forgetting and promote knowledge transfer simultaneously. More importantly, SAPT works under the practical settings where no task-ids are provided for the inputs to select their corresponding parameters. Experimental results also demonstrate the applicability of SAPT across different parameter-efficient tuning methods, models of varying scales and architectures, highlighting its universality.

\section{Limitations}
There are several limitations to consider for future directions of continual learning of large language models. Firstly, when the learning sequence scales to hundreds of tasks, continually expanding the PET pool to allocate a PET block for each one of them would lead to large computation and storage costs. Thus, how to prune and merge similar PET blocks in the continual learning process can be an interesting direction to explore. Secondly, although SAPT exhibits the best performance of Backward Transfer (BWT), it still fails to allow subsequent tasks to impose the positive impacts on the learned ones. This could be a critical direction to further explore more advanced CL methods.  Finally, even though our approach do not depend on identifying task-ids during the testing phase, it still necessitates the identification of tasks during training to establish distinct PET parameters for each task. Investigating techniques for training that is independent of task identification could prove to be a promising avenue for future research, which could favor the application of continual learning upon on the online streams of data.

\section*{Acknowledgements}
We thank the anonymous reviewers for their comments and suggestions. And we thank Libo Qin for the suggestions on the writing of this work.

\bibliography{custom}

\newpage

\appendix

\section{Parameter-Efficient Tuning Methods}
\label{pet}
We adopt two representative PET methods, Prompt Tuning \citep{lester2021power} and LoRA \citep{hu2021lora} in our proposed SAPT, which are referred to as PET blocks in this study.

In prompt tuning, a series of virtual tokens, called soft prompt $P$ is prepended to the input text $x$, while keeping the LLM parameters frozen. In this case, during the training on the downstream tasks, gradient updates are preformed on the prompt parameters independently.

In LoRA, the pre-trained weight matrix of LLMs is updated with a low-rank decomposition. For a linear layer $h=W_0x$, the forward pass with LoRA is modified to be:
\begin{equation}
    h=W_0x+BAx
\end{equation}
where $W_0 \in \mathbb{R}^{d\times k}$, $B \in \mathbb{R}^{d\times r}$, $A \in \mathbb{R}^{r\times k}$, with the rank $r \ll \min(d,k)$. The pre-trained weight matrix $W_0$ remains fixed during training, while A and B contain trainable parameters.

\section{Dataset Details}
\label{dataset}

\subsection{Datasets}
Table \ref{superni} \& \ref{long-sequence} show details of the datasets we used for our experiments, along with their evaluation metrics. Overall, in SuperNI, we choose 3 tasks from dialogue generation (Dialog) \citep{zhang2018personalizing,zang2020multiwoz,peskov2020takes}, information extraction (IE) \citep{santus2015evalution,nye2018corpus,mostafazadeh2020glucose},  question answering (QA) \citep{dasigi2019quoref,talmor2019commonsenseqa}, summarization (Sum) \citep{narayan2018don,gliwa2019samsum,kim2019abstractive} and sentiment analysis (SA) \citep{socher2013recursive,saravia2018carer}, respectively.

For the Long Sequence benchmark, this includes five tasks from the standard CL benchmark (AG News, Amazon reviews, Yelp reviews, DBpedia and Yahoo Answers) \citep{zhang2015character}, four from GLUE benchmark (MNLI, QQP, RTE, SST2) \citep{wang2018glue}, five from SuperGLUE benchmark (WiC, CB, COPA, MultiRC, BoolQ) \citep{wang2019superglue}, and the IMDB movie reviews dataset \citep{maas2011learning}.

And unseen tasks from the SuperNI benchmark are displayed Table \ref{unseen_task}. They also from the five categories of Dialog \citep{wei2018airdialogue,cho2020grounding,aliannejadi2021building}, IE \citep{ollie-emnlp12,zlabinger2020crowd,radev2020dart}, QA \citep{levy2017zero,zhang2018record,min2020ambigqa}, Sum \citep{henderson2014third,syed2020news,hasan2021xl} and SA \citep{sheng2020investigating,lowphansirikul2020scb}.

\subsection{Task Sequence Orders}

We report 4 different task orders used for our experiments in Table \ref{order}.

\section{Implementation Details}
\label{implementation}

Our experiments are implemented with PyTorch \citep{paszke2019pytorch} and Transformer library \citep{wolf2020transformers}. The T5-Large is trained on a single NVIDIA Tesla A800 GPU and the larger backbones T5-XL, T5-XXL, LLaMA-2-7B and LLaMA-2-13B are performed on 4 NVIDIA Tesla A800 using DeepSpeed repository.

For our prompt-based methods, the length of prompts is set to 10. Following \citet{lester2021power}, they are initialized from sampled vocabulary of the backbone model and trained using the Adafactor optimizer. On the SuperNI benchmark, we train SAPT-Prompt with 100 epochs, the constant learning rate of 3e-2 and the batchsize of 32 per GPU. As for the hyper-parameter $\lambda$ in Equation (6), it functions to balance the share attention in the process of attentive learning for the newest task and that in the process of attentive reflection for previous tasks. The larger $\lambda$ means that the attentive reflection contributes more to assist SALS in recalling the shared attention of previous tasks. However, excessive $\lambda$ can impair attentive learning for the current task, thereby weakening knowledge transfer. Here, $\lambda$ is set to 1, which is the relatively optimal balance of the attentive learning and reflection. The hidden dimension $d_p$ of the query projection layer is 100. On the Long Sequence benchmark, the model is trained for 10 epochs with a hierarchical learning rate, 3e-1 for prompts and 1e-2 for the query projection layer. We always keep the total batchsize to 32. And the $\lambda$ and $d_p$ for order3 and order4 is (1.5, 200) and (1.3, 150), respectively. The attention temperature in Equation (2) is $d \times exp(1)$, where $d$ is the LLM backbone dimension size.

For our LoRA-based methods, we use AdamW optimizer to train the model with the learning rate of 3e-4 for T5-Large, 1e-4 for those larger T5-XL and T5-XXL models, 5e-5 for LLaMA-2-7B and 1e-5 for LLaMA-2-13B. For T5 series, the batch size is set to 32 per GPU. On the SuperNI benchmark, the low rank $r$, $\lambda$ and $d_p$ are 4, 0.5 and 100, while they are set to 8, 0.1 and 100 for the Long Sequence benchmark. For LLaMA-2 family, and the batch size is 32 in total. The low rank $r$, $\lambda$ and $d_p$ are both 4, 2 and 100 for the Superni and Long Sequence benchmarks.   The attention temperature in Equation (2) is $sqrt(d)$, where $d$ is the LLM backbone dimension size.

To obtain pseudo samples for our ARM, the prompt length is 300 and is trained for 80 epochs utilizing Adafactor with learning rate of 0.5. And in LoRA, the low-rank $r$ is 8. We train it with AdamW with the learning rate of 0.001 for 5k steps. Batch size is set to 16 for both methods.

Further, we carefully evaluate the official implementations of all baselines, in order to make the comparison as fair as possible. We strictly follow the hyper-parameter settings in their original code, where the prompt size is all set to 10 (except that for LFPT5 of 300) and the LoRA rank is set to 8. If this could not reach the expected performance, we carry out the hyper-parameter search of the learning rate and batchsize for them. Following \citet{sun2019lamol,qin2021lfpt5}, the volume of replay samples is 0.02 of the original training set for SAPT and all replay baseline methods (Replay and LFPT5). Please refer to Appendix \ref{Pseudo_analysis} for deeper analysis for the volume of pseudo samples. All the methods are evaluated for 3 random runs.

\section{Fine-grained Results for the Main Experiments}
\label{overall results}
\begin{figure*}
\centering
\includegraphics[width=0.9\textwidth]{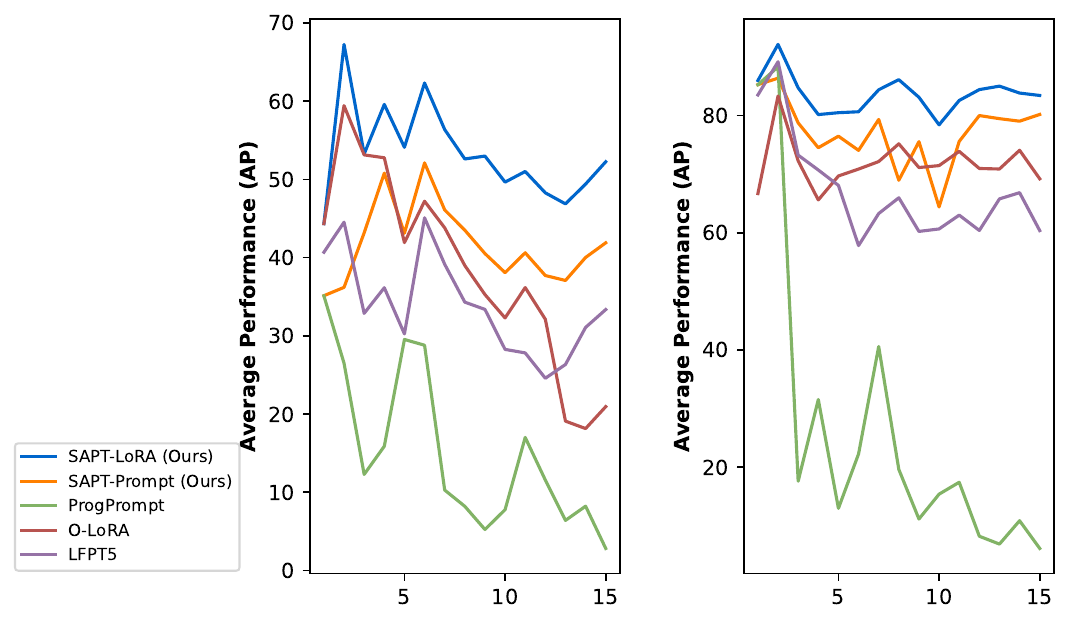}
\caption{The average performance of SAPT and baseline models at each time step on the SuperNI (left) and the Long Sequence (right) benchmark.}
\label{time step}
\end{figure*}

We report the results of each task order on the two benchmark in Table \ref{fin-superni} and Table \ref{fin-long}. And results of the average performance at each time step is displayed in Figure ~\ref{time step}. Overall, the our proposed SAPT demonstrates excellent capabilities in addressing CF and KT.

\section{More Results and Analysis on Generated Pseudo-Samples}

\subsection{Examples of Pseudo Samples}
\label{Pseudo_case}
Table \ref{pseudo_case} shows several pseudo samples generated by SAPT for the SuperNI an Long Sequence Benchmark. Since there are tasks instructions in these two benchmarks, the input-output format of real samples is consists of three elements: [INS] task instruction, [IN] task input and [OUT] task output. And we only generate the input part, [INS] and [IN], to perform attentive reflection in SAPT, which is a novel ways of pseudo-samples usage and greatly different from previous works where complete pseudo samples are generated and mixed with the current task data for multi-task learning. We can see that SAPT can indeed generate high-quality pseudo samples to assist samples from previous tasks in correctly identify the combination of PET blocks specific to each of them.

ARM's efficiency is demonstrated by its need to generate only the input part of samples, unlike previous generative replay methods \citep{sun2019lamol,qin2021lfpt5} that required generating complete (input-output) pairs.

\subsection{Different Volumes and Types of Replayed Samples}
\label{Pseudo_analysis}
\begin{figure*}
\centering
\includegraphics[width=1\textwidth]{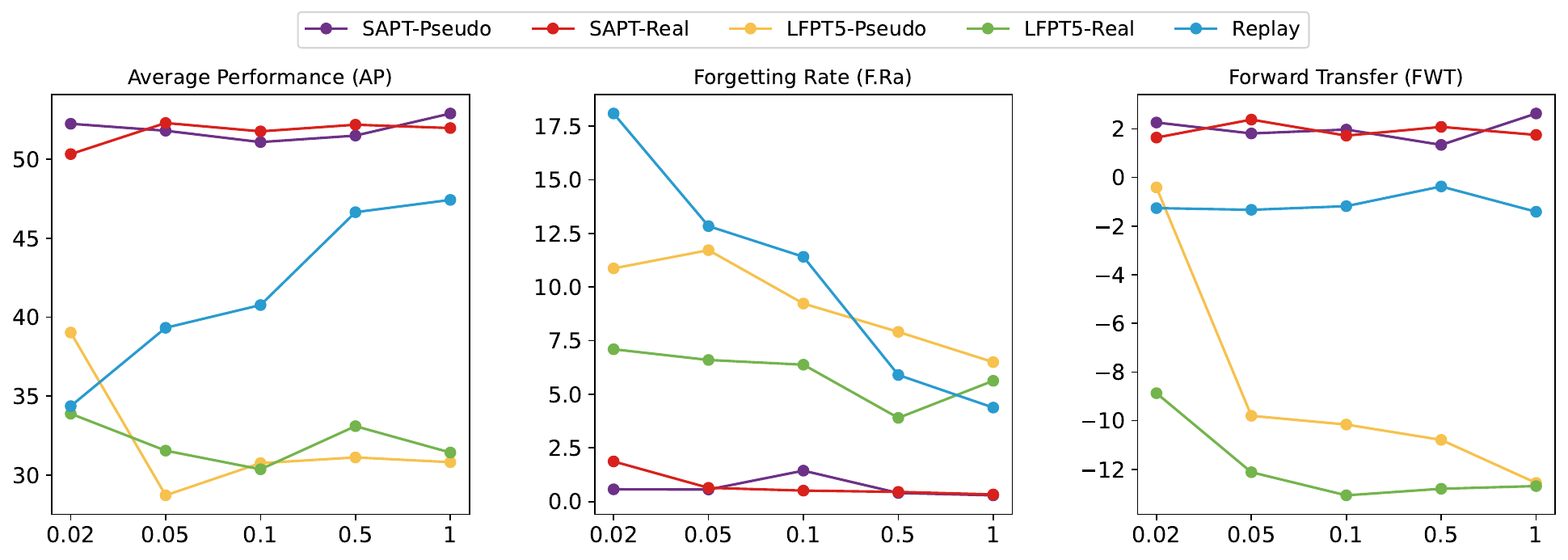}
\caption{Comparison of SAPT-LoRA and baselines based on different types (real and pseudo) and volumes of replayed data, in terms of Average Performance (AP), Forgetting Rate (F.Ra) and Forward Transfer (FWT).}
\label{replay_scale}
\end{figure*}

In SAPT, the Attentive Reflection Module (ARM) provides a novel perspective for utilizing generated pseudo-data. We conduct additional experiments to analyze the impact of using varying scales of pseudo-data and real data on SAPT and the baseline models Replay and LFPT5. The results are shown in Figure \ref{replay_scale}. We have the following two observations that are worth to discuss:

(1) Regardless of whether real data or pseudo-data is used, SAPT demonstrates computational efficiency during replay, showing superior performance even with the minimum replay scale 2\% compared to the maximum replay scale 100\% of LFPT5 and Replay. It is worth mentioning that when the replay data volume of Replay is 100\%, it corresponds to the setting of multi-task learning, which is commonly considered as the upper bound of continual learning. SAPT is able to surpass this upper bound, demonstrating its ability to flexibly handle different inputs, enabling them to be processed by corresponding parameters. 

(2) For SAPT, there is no significant difference in performance between using real data and pseudo-data. This firstly indicates the reliability of the pseudo-data we generated and the sufficient robustness of our proposed ARM, which can utilize pseudo data of different qualities to accomplish reflection on shared attention.

\begin{figure}
\centering
\includegraphics[width=1\columnwidth]{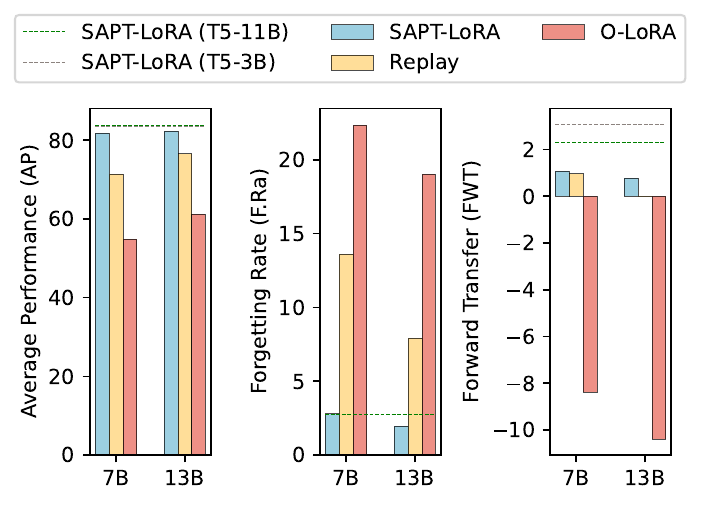}
\caption{Comparison of SAPT and baselines based on different architectures of LLM backbones on the Long Sequence benchmark, including T5 (encoder-decoder) and LLaMA-2 (decoder-only).}
\label{dif_model_long}
\end{figure}

\section{Visualization on Shared Attention}
\label{visual}
\begin{figure*}
\centering
\includegraphics[width=0.9\textwidth]{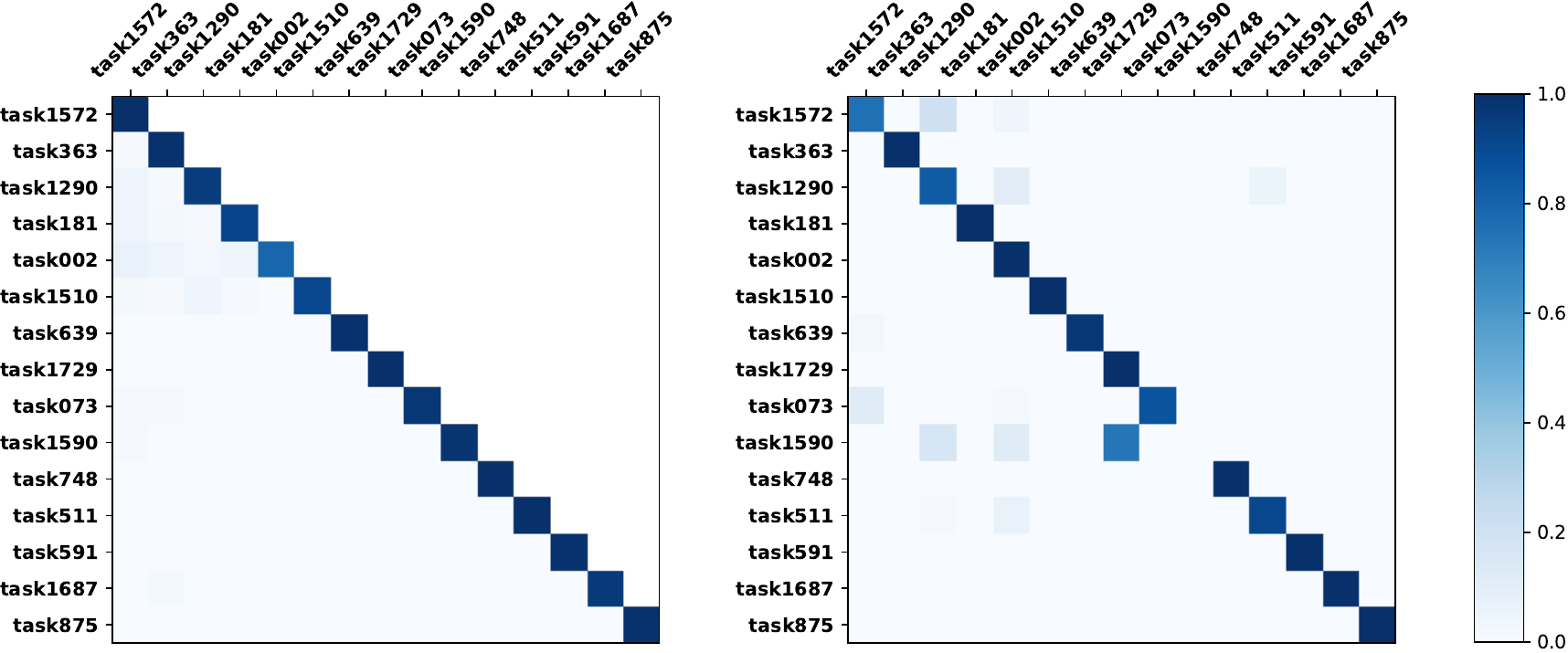}
\caption{Visualization on shared attention of SAPT-Prompt on the SuperNI benchmark during the training (left) and testing time (right).}
\label{vis_prompt_super}
\end{figure*}

\begin{figure*}
\centering
\includegraphics[width=0.9\textwidth]{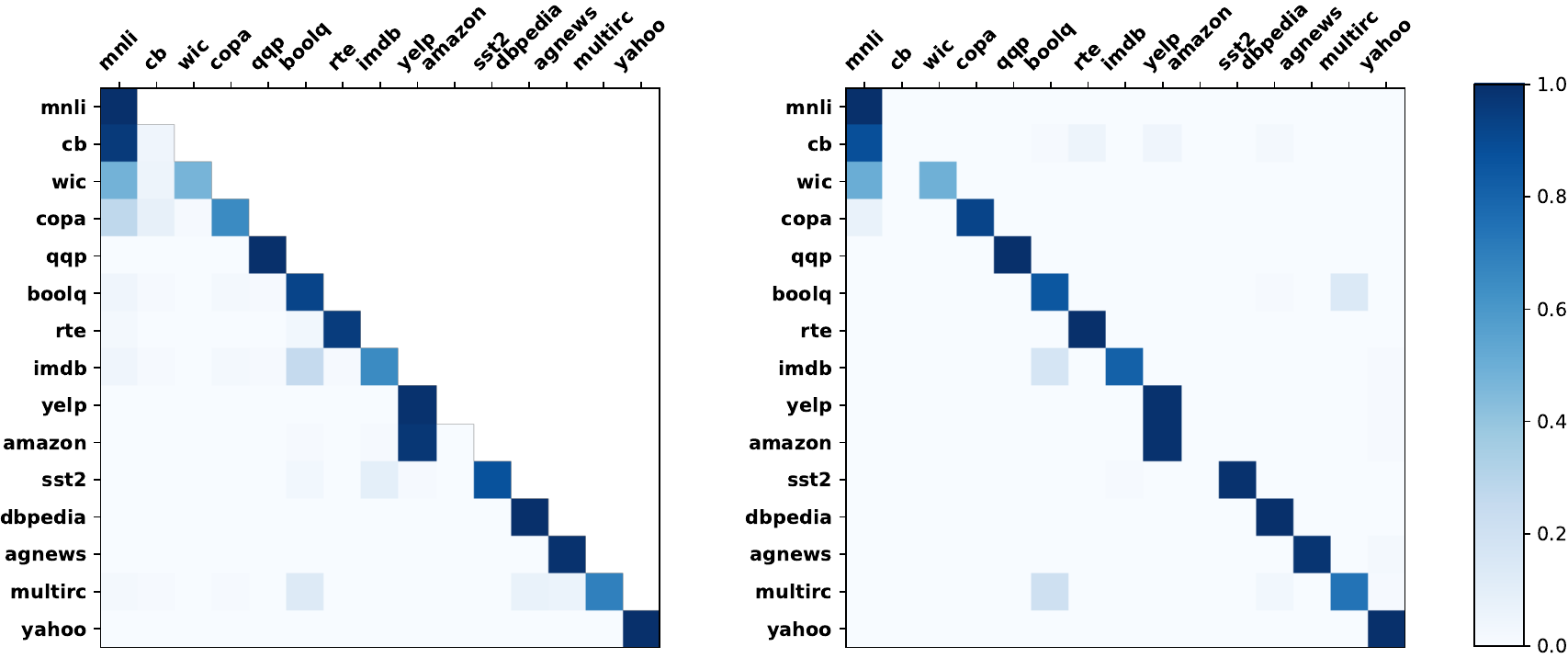}
\caption{Visualization on shared attention of SAPT-Prompt on the Long Sequence benchmark during the training (left) and testing time (right).}
\label{vis_prompt_long}
\end{figure*}

\begin{figure*}
\centering
\includegraphics[width=0.9\textwidth]{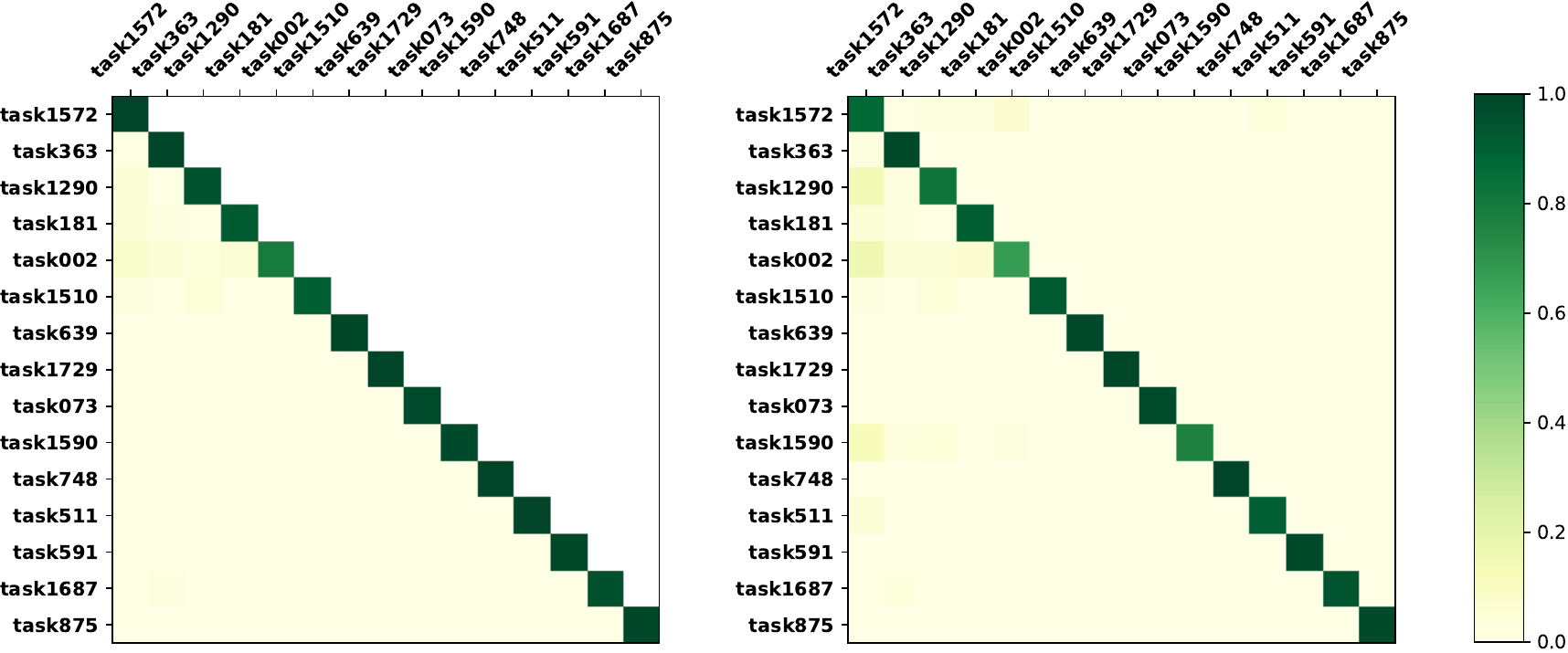}
\caption{Visualization on shared attention of SAPT-LoRA on the SuperNI benchmark during the training (left) and testing time (right).}
\label{vis_lora_super}
\end{figure*}

\begin{figure*}
\centering
\includegraphics[width=0.9\textwidth]{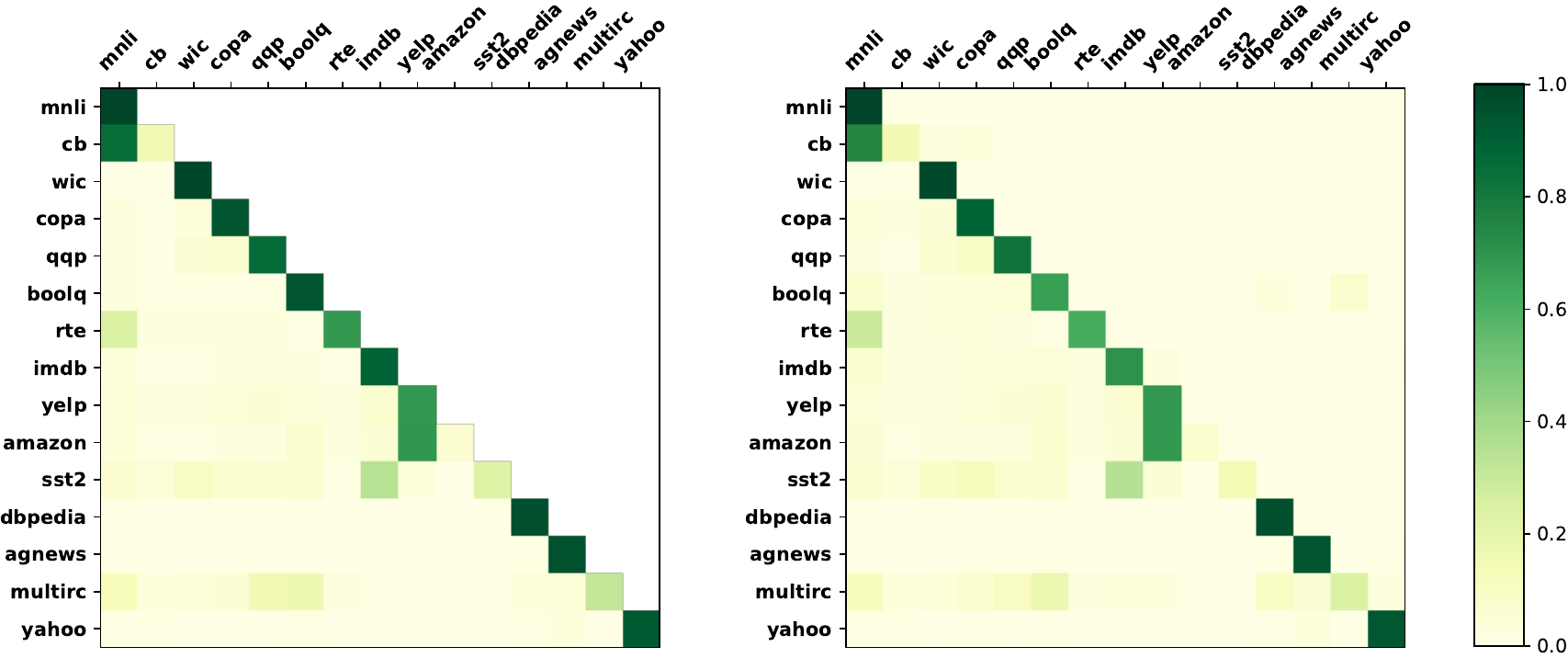}
\caption{Visualization on shared attention of SAPT-LoRA on the Long Sequence benchmark during the training (left) and testing time (right).}
\label{vis_lora_long}
\end{figure*}

We demonstrate the visualization on shared attention operation of SAPT-Prompt on the SuperNI (Figure \ref{vis_prompt_super}) and the Long Sequence (Figure \ref{vis_prompt_long}) Benchmark, and the SAPT-LoRA on the SuperNI (Figure \ref{vis_lora_super}) and the Long Sequence (Figure \ref{vis_lora_long}) Benchmark. And the resulting attention weights is obtained through the average attention weights of the testing samples from a specific task.

Overall, whether based on Prompt or LoRA, SAPT can maintain the alignment for the learning and selection process of PET blocks through shared attention on both benchmarks. Even as the task sequences become longer, it does not affect the ability to identify suitable combinations of PET modules. This directly demonstrates its effectiveness in addressing CF and KT.

Furthermore, both methods demonstrate varying degrees of knowledge transfer on the two benchmarks. Overall, the PET blocks in the current task contribute more significantly, as indicated by the darkest color of the diagonal elements. However, there are also interesting observations where the PET blocks for other tasks have weights higher than the current task, surpassing the higher similarity between these tasks (yelp \& amazon, mnli \& cb). Additionally, the knowledge transfer of Prompt appears slightly more pronounced than LoRA, but overall, LoRA outperforms Prompt in terms of the overall performance. This may be attributed to LoRA's superior representation and learning of task-specific knowledge in the low-rank space, aligning with the conclusions in previous works \citep{hu2021lora,ding2022delta}.

\section{Scale to LLaMA-2 Model}
\label{llama_res}
The results of SAPT and baseline methods on the Long Sequence Benchmark based on different sizes of T5 and LLaMA-2 are shown in Figure \ref{dif_model_long}. It can be observed that our proposed SAPT still exhibits superiority to effectively mitigating CF and promoting KT over baseline methods.

Selecting O-LoRA as the baseline method for experiments based on LLaMA-2 is because it is the only work among numerous baselines that experimented with LLaMA-2 in the original paper, while other baselines are almost originally implemented with T5 or BERT in their paper. Here we additionally supplement the experimental results of EPI, LFPT5 based on LLaMA2-7B and -13B. Results are shown in Table \ref{llama_7b_results} and Table \ref{llama_13b_results}.

\begin{table*}
\centering
\begin{tabular}{lllll}
\toprule
\textbf{Dataset name} & \textbf{Task}  & \textbf{Metric} \\
\midrule
1. task639\_multi\_woz\_user\_utterance\_generation  & dialogue generation   & Rouge-L        \\
2. task1590\_diplomacy\_text\_generation & dialogue generation   & Rouge-L       \\
3. task1729\_personachat\_generate\_next & dialogue generation   & Rouge-L      \\
4. task181\_outcome\_extraction & information extraction & Rouge-L        \\
5. task748\_glucose\_reverse\_cause\_event\_detection & information extraction & Rouge-L       \\
6. task1510\_evalution\_relation\_extraction   & information extraction & Rouge-L  \\
7. task002\_quoref\_answer\_generation & question answering & Rouge-L \\
8. task073\_commonsenseqa\_answer\_generation & question answering & Rouge-L    \\
9. task591\_sciq\_answer\_generation  & question answering & Rouge-L        \\
10. task511\_reddit\_tifu\_long\_text\_summarization     & summarization        & Rouge-L        \\
11. task1290\_xsum\_summarization  & summarization       & Rouge-L        \\
12. task1572\_samsum\_summary  &summarization  & Rouge-L \\
13. task363\_sst2\_polarity\_classification  & sentiment analysis   & accuracy        \\
14. task875\_emotion\_classification & sentiment analysis   & accuracy  \\
15. task1687\_sentiment140\_classification & sentiment analysis   & accuracy  \\
\bottomrule
\end{tabular}
\caption{The details of 15 datasets in the SuperNI Benchmark \citep{wang2022super}.
}
\label{superni}
\end{table*}

\begin{table*}[htbp]
\centering
\begin{tabular}{lllll}
\toprule
\textbf{Dataset name} & \textbf{Category} & \textbf{Task}             & \textbf{Domain}     & \textbf{Metric} \\ \midrule
1. Yelp               & CL Benchmark      & sentiment analysis        & Yelp reviews        & accuracy        \\
2. Amazon             & CL Benchmark      & sentiment analysis        & Amazon reviews      & accuracy        \\
3. DBpedia            & CL Benchmark      & topic classification      & Wikipedia           & accuracy        \\
4. Yahoo              & CL Benchmark      & topic classification      & Yahoo Q\&A          & accuracy        \\
5. AG News            & CL Benchmark      & topic classification      & news                & accuracy        \\
6. MNLI               & GLUE              & natural language
inference                       & various             & accuracy        \\
7. QQP                & GLUE              & paragraph detection       & Quora               & accuracy        \\
8. RTE                & GLUE              & natural language inference                       & news, Wikipedia     & accuracy        \\
9. SST-2              & GLUE              & sentiment analysis        & movie reviews       & accuracy        \\
10. WiC               & SuperGLUE         & word sense disambiguation & lexical databases   & accuracy        \\
11. CB                & SuperGLUE         & natural language
inference                       & various             & accuracy        \\
12. COPA              & SuperGLUE         & question and answering                        & blogs, encyclopedia & accuracy        \\
13. BoolQA            & SuperGLUE         & boolean question and answering                & Wikipedia           & accuracy        \\
14. MultiRC           & SuperGLUE         & question and answering                        & various             & accuracy        \\
15. IMDB              & SuperGLUE         & sentiment analysis        & movie reviews       & accuracy        \\ \bottomrule
\end{tabular}
\caption{The details of 15 classification datasets in the Long Sequence Benchmark \citep{razdaibiedina2022progressive}. First five tasks
correspond to the standard CL benchmark \citep{zhang2015character}.
}
\label{long-sequence}
\end{table*}

\begin{table*}
\centering
\begin{tabular}{lllll}
\toprule
\textbf{Dataset name} & \textbf{Task}  & \textbf{Metric} \\
\midrule
1. task360\_spolin\_yesand\_response\_generation  & dialogue generation   & Rouge-L        \\
2. task574\_air\_dialogue\_sentence\_generation & dialogue generation   & Rouge-L       \\
3. task1714\_convai3\_sentence\_generation & dialogue generation   & Rouge-L      \\
4. task180\_intervention\_extraction & information extraction & Rouge-L        \\
5. task678\_ollie\_actual\_relationship\_answer\_generation & information extraction & Rouge-L       \\
6. task1410\_dart\_relationship\_extraction   & information extraction & Rouge-L  \\
7. task339\_record\_answer\_generation & question answering & Rouge-L \\
8. task670\_ambigqa\_question\_generation & question answering & Rouge-L    \\
9. task1327\_qa\_zre\_answer\_generation\_from\_question  & question answering & Rouge-L        \\
10. task522\_news\_editorial\_summary     & summarization        & Rouge-L        \\
11. task1356\_xlsum\_title\_generation  & summarization       & Rouge-L        \\
12. task1499\_dstc3\_summarization  &summarization  & Rouge-L \\
13. task421\_persent\_sentence\_sentiment\_classification  & sentiment analysis   & accuracy        \\
14. task833\_poem\_sentiment\_classification & sentiment analysis  & accuracy  \\
15. task929\_products\_reviews\_classification & sentiment analysis   & accuracy  \\
\bottomrule
\end{tabular}
\caption{The details of unseen tasks from the SuperNI benchmark.
}
\label{unseen_task}
\end{table*}

\begin{table*}
\centering
\begin{tabular}{lll}
\toprule
\textbf{Order} & \textbf{Model} & \textbf{Task Sequence}                                                                                                                                \\ 
\midrule
1              & T5, LLaMA-2      & \begin{tabular}[c]{@{}l@{}}task1572 → task363 → task1290 → task181 → task002 →\\ task1510 → task639 → task1729 → task073 → task1590 →\\ task748 → task511 → task591 → task1687 → task875\end{tabular} \\
2              & T5, LLaMA-2     & \begin{tabular}[c]{@{}l@{}}task748 → task073 → task1590 → task639 → task1572 →\\ task1687 → task591 → task363 → task1510 → task1729 →\\ task181 → task511 → task002 → task1290 → task875\end{tabular} \\
\midrule
3              & T5, LLaMA-2             & \begin{tabular}[c]{@{}l@{}}mnli → cb → wic → copa → qqp → boolqa → rte → imdb →\\ yelp → amazon → sst-2 → dbpedia → ag → multirc → yahoo\end{tabular} \\
4              & T5, LLaMA-2             & \begin{tabular}[c]{@{}l@{}}yelp → amazon → mnli → cb → copa → qqp → rte → imdb →\\ sst-2 → dbpedia → ag → yahoo → multirc → boolqa → wic\end{tabular} \\
\bottomrule
\end{tabular}
\caption{Four different orders of task sequences used for our experiments. Orders
1-2 correspond to the SuperNI benchmark. Orders 3-4 are long-sequence orders following \citet{razdaibiedina2022progressive}.}
\label{order}
\end{table*}

\begin{table*}[]
\small
\centering
\begin{tabular}{
l |
c 
c 
c 
c |
c 
c 
c 
c }
\toprule

\textbf{}        & \multicolumn{4}{c|}{\textbf{Order 1}} & \multicolumn{4}{c}{\textbf{Order 2}} \\
\textbf{}        & \textbf{AP}$\uparrow$      & \textbf{F.Ra}$\downarrow$      & \textbf{FWT}$\uparrow$      & \textbf{BWT}$\uparrow$      & \textbf{AP}$\uparrow$        & \textbf{F.Ra}$\downarrow$       & \textbf{FWT}$\uparrow$       & \textbf{BWT}$\uparrow$     \\ \midrule
SeqLoRA  &  5.05                &  30.94           &   -17.01        &  -28.88     &7.80          &   35.84       &  -10.15     & -32.99         \\
Replay   & 34.37         &  18.09      & -1.26    & -16.89     &  36.37       &  15.74      & -1.44     & -14.69       \\
L2P    &  15.18     &  6.23     &  -20.97       &  -3.65        &  10.27        &  17.51   & -17.30    & -12.24        \\
LFPT5  & 39.03 & 10.87  & -0.41 & -9.85   & 29.70 & 20.72 & -0.51 & -19.08 \\
ProgPrompt & 2.83   &35.65    &-3.70   &-33.27  & 3.85  & 35.48   & -2.87  & -33.09   \\
EPI  & -               & -            & -            & -              & -               & -            & -            & -             \\
O-LoRA  & 20.95               & 30.91            & -0.43            & -28.83              & 30.82              & 21.83             & 0.15             & -20.35             \\ \midrule
\textbf{SAPT-Prompt}     & 41.88              & 1.41              & \textbf{2.83}              & -0.75                & 40.34              & \textbf{1.23}             & 1.07             & \textbf{-0.54}             \\
\textbf{SAPT-LoRA}   & \textbf{52.25}               & \textbf{0.57}              & 2.26              & \textbf{-0.23}                & \textbf{50.82}              & 1.24             & \textbf{1.50}             & -0.90             \\ 
\bottomrule
\end{tabular}
\caption{The overall results on each task order of the SuperNI benchmark with T5-Large model. Performance of continual learning (AP), forgetting rate (F.Ra), forward transfer (FWT) and backward transfer (BWT) are reported after training on the last task.}
\label{fin-superni}
\end{table*}

\begin{table*}[]
\small
\centering
\begin{tabular}{
l |
c 
c 
c 
c |
c 
c 
c 
c }
\toprule

\textbf{}        & \multicolumn{4}{c|}{\textbf{Order 3}} & \multicolumn{4}{c}{\textbf{Order 4}} \\
\textbf{}        & \textbf{AP}$\uparrow$      & \textbf{F.Ra}$\downarrow$      & \textbf{FWT}$\uparrow$      & \textbf{BWT}$\uparrow$      & \textbf{AP}$\uparrow$        & \textbf{F.Ra}$\downarrow$       & \textbf{FWT}$\uparrow$       & \textbf{BWT}$\uparrow$     \\ \midrule
SeqLoRA  &  6.71                &  82.07           &   1.19        &  -76.60     & 12.73    & 75.15  & 0.43  & -70.14        \\
Replay   &  68.20  & 16.21 &1.20  &  -15.13    & 74.25      & 9.89     & 1.36   & -9.23      \\
L2P    & 58.61  &   21.55    &   1.01      &   -15.43 &   57.34   & 23.42 &   1.70        & -17.82        \\
LFPT5 & 66.62 & 14.57 & 2.89 & -13.60   & 67.40        &   13.20          &   2.06        &   -11.99       \\
ProgPrompt       & 6.14  &  74.64  & -1.65  & -69.53 & 9.83  & 68.45   & -3.61  & -63.89  \\
EPI  &75.19 & 0.77 & -1.54 & \textbf{-0.60}   & 75.10& 2.44& 0.00 &-2.23            \\
O-LoRA  & 69.22  & 8.30  & -7.79   & -4.42    & 69.26              & 5.70             & -8.51 & -5.09             \\ \midrule
\textbf{SAPT-Prompt}     & 80.20              & 0.91              & \textbf{3.63}              & -0.76                & 78.08              & 2.45             & \textbf{2.95}             & -2.20             \\
\textbf{SAPT-LoRA}   & \textbf{83.44}               & \textbf{0.75}              &1.99              & -0.66                & \textbf{80.60}              & \textbf{2.25}             & 1.72             & \textbf{-1.94}             \\ 
\bottomrule
\end{tabular}
\caption{The overall results on each task order of the Long Sequence benchmark with T5-Large model. Performance of continual learning (AP), forgetting rate (F.Ra), forward transfer (FWT) and backward transfer (BWT) are reported after training on the last task.}
\label{fin-long}
\end{table*}

\begin{table*}[h!]
\small
\centering
\resizebox{0.8\linewidth}{!}{
\begin{tabular}{l |c c c c |c c c c }
\toprule

\textbf{}        & \multicolumn{4}{c|}{\textbf{SuperNI Benchmark}} & \multicolumn{4}{c}{\textbf{Long Sequence Benchmark}} \\
\textbf{}        & \textbf{AP}$\uparrow$      & \textbf{F.Ra}$\downarrow$      & \textbf{FWT}$\uparrow$      & \textbf{BWT}$\uparrow$      & \textbf{AP}$\uparrow$        & \textbf{F.Ra}$\downarrow$       & \textbf{FWT}$\uparrow$       & \textbf{BWT}$\uparrow$     \\ \midrule
Replay  &  39.48	&14.86	&0.19	&-26.47	&71.43	&13.64	&0.97	&-12.73     \\
LFPT5 & 38.71	&16.81	&0.32	&-15.42	&70.31	&5.63	&0.51	&-4.32      \\
EPI  & -               & -            & -            & -              & 72.27	&5.04	&-3.12	&\textbf{-0.50}             \\
O-LoRA  & 24.26	&27.56	&-6.09	&-25.73	&54.95	&22.36	&-8.38	&-19.86             \\ \midrule
\textbf{SAPT-Prompt}	&47.39	&2.12	&\textbf{0.92}	&-2.02	&77.62	&3.29	&0.33	&-2.98 \\
\textbf{SAPT-LoRA}   & \textbf{56.23}	&\textbf{1.07}	&0.81	&\textbf{-0.65}	&\textbf{81.75}	&\textbf{2.81}	&\textbf{1.09}	&-2.53
\\ 
\bottomrule
\end{tabular}
}
\caption{The overall results on two continual learning benchmarks with LLaMA-2-7B model. Performance of continual learning (AP), forgetting rate (F.Ra), forward transfer (FWT) and backward transfer (BWT) are reported after training on the last task. All results are averaged over two different orders of each benchmark.}
\label{llama_7b_results}
\end{table*}

\begin{table*}[h!]
\small
\centering
\resizebox{0.8\linewidth}{!}{
\begin{tabular}{l |c c c c |c c c c }
\toprule

\textbf{}        & \multicolumn{4}{c|}{\textbf{SuperNI Benchmark}} & \multicolumn{4}{c}{\textbf{Long Sequence Benchmark}} \\
\textbf{}        & \textbf{AP}$\uparrow$      & \textbf{F.Ra}$\downarrow$      & \textbf{FWT}$\uparrow$      & \textbf{BWT}$\uparrow$      & \textbf{AP}$\uparrow$        & \textbf{F.Ra}$\downarrow$       & \textbf{FWT}$\uparrow$       & \textbf{BWT}$\uparrow$     \\ \midrule
Replay  &  43.99	&11.64	&0.72	&-9.75	&76.63	&7.92	&0.02	&-14.86     \\
LFPT5 & 41.26	&14.67	&-0.52	&-12.31	&71.61	&6.51	&-1.34	&-3.78  \\
EPI  & -               & -            & -            & -              & 76.66	&4.91	&-0.09	& \textbf{-1.03}\\
O-LoRA  & 31.18	&24.26	&-4.05	&-22.64	&61.21	&19.03	&-10.4	&-17.54             \\ \midrule
\textbf{SAPT-Prompt}     & 52.31	&1.57	&\textbf{1.49}	&-1.41	&78.54	&3.26	&0.14	&-2.98 \\
\textbf{SAPT-LoRA}   & \textbf{56.95}	&\textbf{1.39}	&0.81	&\textbf{-0.56}	&\textbf{82.32}	&\textbf{1.98}	&\textbf{0.78}	&-1.57
\\ 
\bottomrule
\end{tabular}
}
\caption{The overall results on two continual learning benchmarks with LLaMA-2-13B model. Performance of continual learning (AP), forgetting rate (F.Ra), forward transfer (FWT) and backward transfer (BWT) are reported after training on the last task. All results are averaged over two different orders of each benchmark.}
\label{llama_13b_results}
\end{table*}

\vspace{-5cm}

\begin{table*}[h]
\tiny
    \centering
    \begin{tabular}{l|l|l|p{0.64\textwidth}}
    \hline
    \textbf{Benchmark} & \textbf{Task Name} & \textbf{Type} & \textbf{Data Sample} \\
    \hline
         \multirow{39}{*}{\centering SuperNI} 
         & \multirow{21}{*}{task002\_quoref\_answer\_generation} & \multirow{12}{*}{Real} & [INS] In this task, you're expected to write answers to questions involving multiple references to the same entity. The answer to the question should be unambiguous and a phrase in the paragraph. Most questions can have only one correct answer. [IN] Passage: Phaedra is a poor Greek sponge diver on the island of Hydra. She works from the boat of her boyfriend, Rhif, an illegal immigrant from Albania. She accidentally finds an ancient Greek statue of a boy riding a dolphin on the bottom of the Aegean Sea. Her efforts to sell it to the highest bidder lead her to two competing individuals: Dr. James Calder, an honest archaeologist who will surrender it to Greek authorities, and Victor Parmalee, an aesthete and an unscrupulous dealer with a history of trying to acquire works of art stolen by the Nazis from their owners.\textbackslash nCalder and Parmalee each try to win Phaedra's cooperation. She works in concert with Parmalee, while developing feelings for Calder. When she seems to waver, Rhif decides to make the deal with Parmalee work. The film reaches a happy conclusion, with virtue rewarded, the statue celebrated by the people of Hydra, and Phaedra and Calder in each other's arms. Parmalee, a man with no apparent national loyalties or heritage, sets course for Monte Carlo. \textbackslash nQuestion: Who is dating an illegal immigrant? [OUT] Phaedra. \\
         \cline{3-4}
         & \multirow{9}{*} & \multirow{9}{*}{Pesudo} & [INS] In this task, you're expected to write answers for questions involving multiple references to the same entity. The answer to the question should be unambiguous and a phrase in the paragraph. Most questions can have only one correct answer. [IN] Passage: In 1899 British President Johnston (now known as John Marshall) is charged with running riots against officers of the Royal Navy. After two days of debriefing, Admiral Sir William Rogers Jr. becomes interested in Robert \"Rob\" Johnson, another officer on staff, who finds out that Johnson was not present at the rehearsals for the film; however, his father also claims that he has been fired from the navy because Mr Williams had been shot before him. He then went off to see an officer looking after the military. He later admitted himself guilty—for failing which he became involved in the crime? Although no arrest was made during his service they served less than 40 days. He does remain behind but never gets away or die first. He takes control over \\
         \cline{2-4}
         & \multirow{6}{*}{task875\_emotion\_classification} & \multirow{3}{*}{Real} & [INS] In this task, you are given a sentence containing a particular emotion. You must classify the sentence into one of the six emotions: 'joy', 'love', 'anger', 'fear', or 'surprise'. [IN] i could clearly feel my adomen muscles contract everytime i cough like some adomen exercise haha and im aching from it now sigh [OUT] sadness \\
         \cline{3-4}
         & \multirow{3}{*} & \multirow{3}{*}{Pesudo} & [INS] In this task, you are given a sentence containing monetary hardship. You must classify the sentence into one of the six emotions: 'joy', 'love', 'anger', 'fear', or'surprise'. [IN] i feel like it is worthwhile to support local artists and so does clay so im amazed by how supportive and passionate artistic writers can be \\
         \cline{2-4}
         & \multirow{12}{*}{task1572\_samsum\_summary} & \multirow{5}{*}{Real} & [INS] In this task, you are given a conversation, and your task is to generate a summary from the information present in the given conversation. Generate a summary in such a way that the context should be present in the conversation. It should cover the complete context of the conversation. [IN]  Mary: Can you pick up the kids?, John: Sure. You've already asked me this morning., Mary: Really? I don't remember. , John: Ok, doesn't matter.  [OUT]  John will pick up the kids at Mary's request.  \\
         \cline{3-4}
         & \multirow{7}{*} & \multirow{7}{*}{Pesudo} & [INS] In this task, you are given a conversation, and your task is to generate a summary from the information present in the given conversation. Generate a summary in such a way that the context should be present in the conversation. It should cover the complete context of the conversation. [IN] Leda: Ok, so first Homo Faber., Gaby: I've already booked the tickets., Ivan: Where exactly?, Leda: Isola San Giorgio., Gaby: They have limited spaces., Ivan: They are always full, but we can ask around it., Gaby: So for us? We won't make it all day., Marc: OK, what about Giardini or Arsenale? I think we won't make it to see both. We can ask around., Gaby: I'll go back!, Ivan: Me too. We could also go different places... There are a lot of satellitanni: What about the transportation? Same ticket \\
    \cline{1-4}
         \multirow{27}{*}{\centering Long Sequence} 
         & \multirow{9}{*}{amazon} & \multirow{4}{*}{Real} & [INS] Given a review text from an e-commerce platform, please analyze the sentiment polarity expressed in the text. Choose one from the following sentiment categories: very negative, negative, neutral, positive and very positive.\textbackslash n[IN] Title: Great\textbackslash nText: Jim Cramer is the best - his stock tips and philosophy of investing is truly outstanding.Thanks Jim for all of your insight!\textbackslash n[OUT] positive \\
         \cline{3-4}
         & \multirow{5}{*} & \multirow{5}{*}{Pesudo} & [INS] Given a review text from an e-commerce platform, please analyze the sentiment polarity expressed in the text. Choose one from the following sentiment categories: very negative, negative, neutral, positive and very positive! [IN] Title: Just not as cool as I thought it would be Text: This was just too cool to give up on this movie! Unfortunately, there are way more people with bad taste than Meg McConnell here or even Amy Schumer who make some of her own movies instead of being a good mommy. \\
         \cline{2-4}
         & \multirow{13}{*}{imdb} & \multirow{5}{*}{Real} & [INS] Classify the sentiment polarity expressed in the following movie reviews.\textbackslash n [IN] The Japanese \"Run Lola Run,\" his is one offbeat movie which will put a smile on just about anyone's face. Fans of Run Lola Run, Tampopo, Go!, and Slacker will probably like this one. It does tend to follow a formula that is increasingly popular these days of separate, seemingly unrelated vignettes, all contributing the the overall story in unexpected ways. catch it if you see it, otherwise wait for the rental. [OUT] Good \\
         \cline{3-4}
         & \multirow{8}{*} & \multirow{8}{*}{Pesudo} & [INS] Classify the sentiment polarity expressed in this movie reviews. [IN] Well, I've just got back from watching this film for my birthday and it's been an incredible experience to see how well done that film is! The production values are unbelievable; not only does it look really good, but they have amazing acting: John Wayne makes one of the best actors on his career, and when he plays Jack Ryan (Scott Stierlin) there can be very few performances you would want him to lose weight because he is over 50 pounds. It should all appeal to everybody. There must also explain why such great cinema was important at the time so many years ago. This picture is another highlight which shows how bad James Bond could be like with anything else. We know what we were facing before leaving our country after 24 hours or 48 days...and now we have had some real talent in this film!! \\
         \cline{2-4}
         & \multirow{4}{*}{yahoo} & \multirow{2}{*}{Real} & [INS] I will give you a pair of question and answer, please categorize the topic discussed within.\textbackslash n [IN]  Question: who is the actress from india at da vinci premiere?\textbackslash nAnswer: Aishwarya Rai, the former Miss World.\textbackslash n [OUT] Entertainment \& Music \\
         \cline{3-4}
         & \multirow{2}{*} & \multirow{2}{*}{Pesudo} & [INS] I will give you a pair of question and answer, please categorize the topic discussed within. [IN] Question: what is the word "butterfly"? Answer: It means butterfly-like flower that grow in a basket or vase with lilies on it \\
             
    \hline
    \end{tabular}
\caption{Examples of generated pseudo samples of the SuperNI and the Long Sequence benchmarks. [INS], [IN] and [OUT] represent the task instruction, task input and task output, respectively.}
\label{pseudo_case}
\end{table*}

\end{document}